\documentclass[review]{elsarticle}
\pdfoutput=1
\usepackage{lineno,hyperref}
\modulolinenumbers[5]
\usepackage{amsmath}
\usepackage{amssymb}
\usepackage{latexsym}
\usepackage{mathrsfs}
\usepackage{graphicx}
\usepackage{graphics}
\usepackage{epstopdf,xcolor,listings}
\usepackage{diagbox,lscape}
\usepackage{microtype,enumerate}
\usepackage{booktabs} 
\usepackage{makecell,diagbox}
\usepackage{siunitx}

\usepackage{cleveref}
\usepackage{ulem}
\usepackage{threeparttable}
\usepackage{subcaption}
\usepackage{appendix}
\usepackage{color}
\usepackage{ulem}
\usepackage{cancel,soul}
\usepackage{cleveref}
\usepackage{tikz}
\usepackage{booktabs}
\newcommand*{\circled}[1]{\lower.7ex\hbox{\tikz\draw (0pt, 0pt)%
		circle (.5em) node {\makebox[1em][c]{\small #1}};}}
\usepackage{algorithm}
\usepackage{algorithmic}
\renewcommand{\algorithmicrequire}{\textbf{Input:}}
\renewcommand{\algorithmicensure}{\textbf{Output:}}
\usepackage{bm}
\usepackage[numbers]{natbib}
\usepackage{graphicx,color,multirow,amsmath}
\usepackage{indentfirst}
\usepackage{amsthm}
\usepackage{mathrsfs}
\usepackage{caption}
\newtheorem{remark}{Remark}
\definecolor{mygray}{rgb}{0.92,0.92,0.92}

\journal{Automation in Construction}

\usepackage{changes}

\begin{document}

\begin{frontmatter}
	
\title{AL-iGAN: An Active Learning Framework for Tunnel Geological Reconstruction Based on TBM Operational Data}

\author[mymainaddress]{Hao~Wang}
\author[mymainaddress]{Lixue~Liu}
\author[mysecondaryaddress]{Xueguan~Song}
	
\author[mymainaddress]{Chao~Zhang\corref{mycorrespondingauthor}}
\cortext[mycorrespondingauthor]{Corresponding author}
\ead{chao.zhang@dlut.edu.cn}

\author[mythirdaddress]{Dacheng~Tao}
	
\address[mymainaddress]{School of Mathematical Sciences, Dalian University of Technology, Dalian, 116024, China}
	
\address[mysecondaryaddress]{School of Mechanical Engineering, Dalian University of Technology, Dalian, 116024, China}

\address[mythirdaddress]{JD Explore Academy, JD.COM Inc., 100176, China}

\begin{abstract}

In tunnel boring machine (TBM) underground projects, an accurate description of the rock-soil types distributed in the tunnel can decrease the construction risk ({\it e.g.} surface settlement and landslide) and improve the efficiency of construction. In this paper, we propose an active learning framework, called AL-iGAN, for tunnel geological reconstruction based on TBM operational data. This framework contains two main parts: one is the usage of active learning techniques for recommending new drilling locations to label the TBM operational data and then to form new training samples; and the other is an incremental generative adversarial network for geological reconstruction (iGAN-GR), whose weights can be incrementally updated to improve the reconstruction performance by using the new samples. The numerical experiment validate the effectiveness of the proposed framework as well.

\end{abstract}


\end{frontmatter}


\section{Introduction}
\label{sec:introduction}

With the advantage of high advance rate, environmental friendliness and security, tunnel boring machines (TBMs) have taken the place of the traditional tunneling techniques ({\it e.g.}, drilling and blasting) as the primary construction method for various underground excavation projects such as highways \cite{liao2008analysis}, railway \cite{jin2018analysis} and water conservancy \cite{yagiz2015application}. However, there are still many challenges in the project applications of TBMs because it is difficult to explore the geological information in the TBM construction tunnel. Especially in some complex geological conditions, such as fracture zones, geological faults, aquifers, karst caves and underground rivers, the TBM excavation process will come with a high risk caused by an unanticipated presence of water leakage, caves and hard rocks, and even the finished TBM tunnel will also be faced with the risk of surface settlement and landslide. For instance, there have arisen two accidents of ground collapse caused by water and sand gushing during TBM construction in Nanjing and Tianjing, two Chinese cites, in 2007 and 2011, respectively. Therefore, an accurate description of the rock-soil types distributed in the TBM construction tunnel can effectively reduce the construction risk and improve the construction efficiency \cite{cheng2020identifying}.

\subsection{Background and Motivation}

\begin{figure}[htbp]
	\centering
	\begin{subfigure}[b]{1\textwidth}
		\centering
		\includegraphics[width=\textwidth]{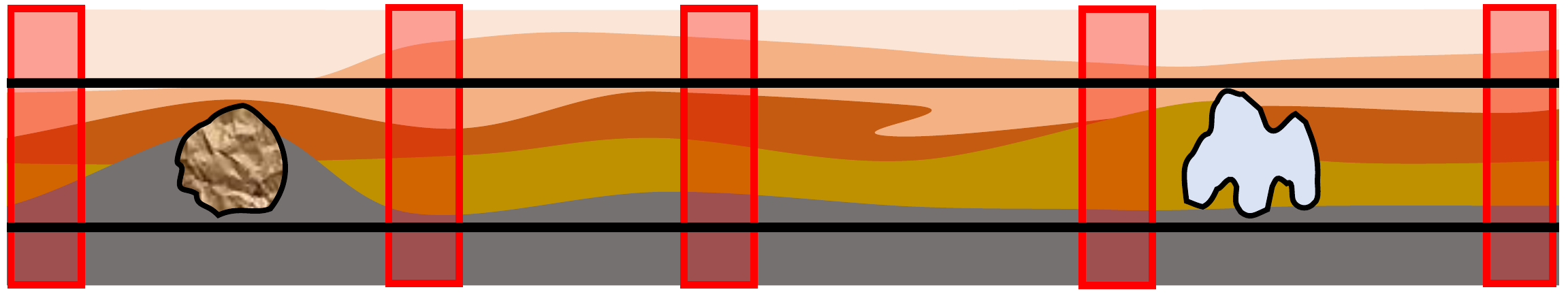}
		\caption{Real rock-soil distribution and discrete sampling locations (marked in red color)}
	\end{subfigure}

	\begin{subfigure}[b]{1\textwidth}
		\centering
		\includegraphics[width=\textwidth]{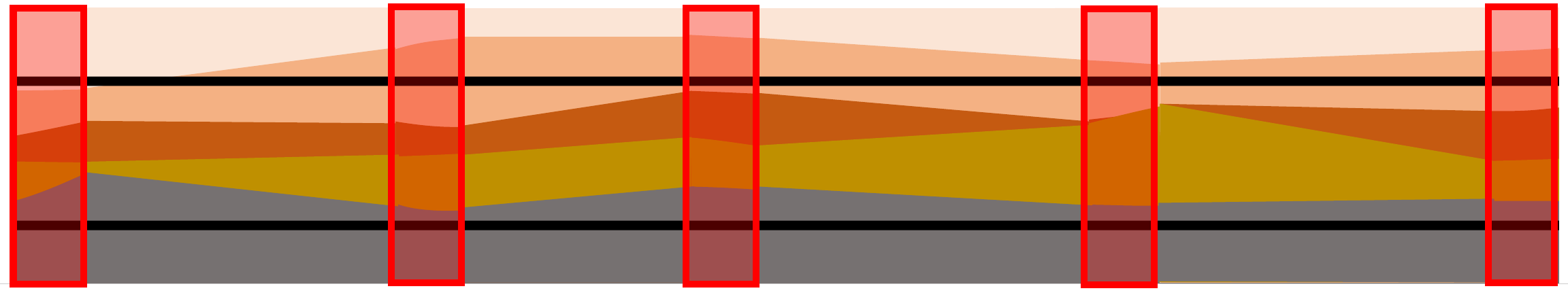}
		\caption{Geological reconstruction result obtained by using interpolation methods based on the discrete geological samples}
	\end{subfigure}
	\caption{Geological Reconstruction Based on Discrete Geological Sampling}
	\label{fig:reconstruction}
\end{figure}

The tunnel geological reconstruction is a principal means for exploring the geological information of the entire TBM tunnel. Compared with the geological prediction, which produces the estimation to the geological condition ahead of the tunnel face, the geological reconstruction aims to accurately describe the distribution information of rock-soil types appearing in the entire tunnel. A number of approaches have been developed to understand the geological condition, including the measurement-based methods and the data-based methods. The measurement-based methods adopt the destructive methods ({\it e.g.} drilling and excavating \cite{chen2017geostatistical, park2017predicting}) or the non-destructive methods ({\it e.g.} the electrode equipped on the tunnel face \cite{park2016predicting} and the ground penetrating radar (GPR) \cite{liu2018forward}) to directly detect the rock-soil distribution information in a few specific locations along the tunnel alignment. However, it is almost technically impossible to accurately reconstruct the rock-soil distribution throughout the entire tunnel by only using the geological information at the discrete locations ({\it cf.} Fig. \ref{fig:reconstruction}).

As addressed above, since the measurement-based methods cannot directly detect the geological condition of the entire tunnel, the reconstruction results are likely to omit some crucial geological information. Instead, a number of different kinds of sensors are equipped on the TBM to collect its real-time operational data during the tunneling process, and the geological information of the entire construction tunnel will be encoded into the collected TBM operational data. The data-based geological reconstruction methods decode the useful features from TBM operational data indexed by the continuous time or displacement points, and then to develop the learning models for estimating the rock-soil distribution of the entire tunnel. The main research concerns on the data-based methods lie in the following three aspects: 

\begin{itemize}
\item feature extraction --- how to extract useful features from the TBM operational data that are indexed by the continuous time or displacement;

\item learning model --- how to design a reasonable regressor for estimating the thickness of each rock-soil type appearing in the tunnel alignment.

\item quantity of data --- how to handle the imbalance between the TBM operational data labeled with the geological information and the unlabeled TBM operational data;

\end{itemize}

In view of the sequentiality of TBM operational data, Liu {\it et al.} \cite{liu2021hard} employed the long-short-term-memory networks to capture the sequential characteristics of the training data and then to predict rock-soil types at the tunnel face. Leu and Adi \cite{leu2011probabilistic} presented a model based on Hidden Markov model and neural network sand predict the probabilistic description of rock-soil types. Zhang {\it et al.}\cite{zhang2019prediction} utilized hierarchies algorithm to compressed data and adopted support vector classifiers as the learning models for justifying the appearance of some specific rock-soil types (rather than estimating their thickness). Zhao {\it et al.}\cite{zhao2019data} augmented features by using the first-order and the second-order difference information and designed a feed-forward artificial neural network (ANN) with two hidden layers as the predictor, where the second hidden layer has $ 7 $ nodes that correspond to $ 7 $ kinds of physical-mechanical indexes. Zhang \cite{zhang2022generative} designed the generative adversarial networks for predicting the thicknesses of each rock-soil types based on TBM operational data, where the adversarial training strategy was adopted to exhaustively overcome the negative effects of the rarity of labeled TBM operational data. One main difficulty of these data-based methods lies in the quantity of data. Especially, when the labeled data are very few, these methods are no longer be available. Although one can supplement the labeled data by collecting the geological information at new locations, it is usually expensive and time-consuming to use the geological-condition detection methods ({\it e.g.} drilling or GPR). A compromise way to handle this issue is to recommend new geological-condition detection locations to enrich the labeled operational data based on the training behavior of the reconstruction model developed by using the current labeled and unlabeled data.

\subsection{Active Learning}

In the traditional learning tasks, the samples are passively imposed into the learner in the training phase, and thus this kind of learning processes are also called passive learning. In contrast, the active learning refers to the learning process that the learner can actively explore the informative samples from the sample space and then add them into the subsequent training phase ({\it cf.} Fig. \ref{fig:al}). In this manner, the learning models developed by the active learning methods need a smaller size of labeled data than the learning models derived from the passive learning \cite{ren2020survey}. Taking advantage of this character, the active learning has been widely used in various applications, such as image recognition \cite{gal2017deep}, text classification \cite{goudjil2018novel}, visual question answering \cite{lin2017active}, and object detection \cite{brust2018active}. Kim \textit{et al.} \cite{kim2020towards} proposed a deep active learning approach to effectively reduce the number of training samples in the task of the vision-based monitoring on construction sites. In general, according to the query strategy ({\it i.e.,} the means of selecting the candidate data from the data pool), the active learning methods can be sorted into two types:

\begin{figure}[htbp]
	\centering
	\includegraphics[width=4.5in]{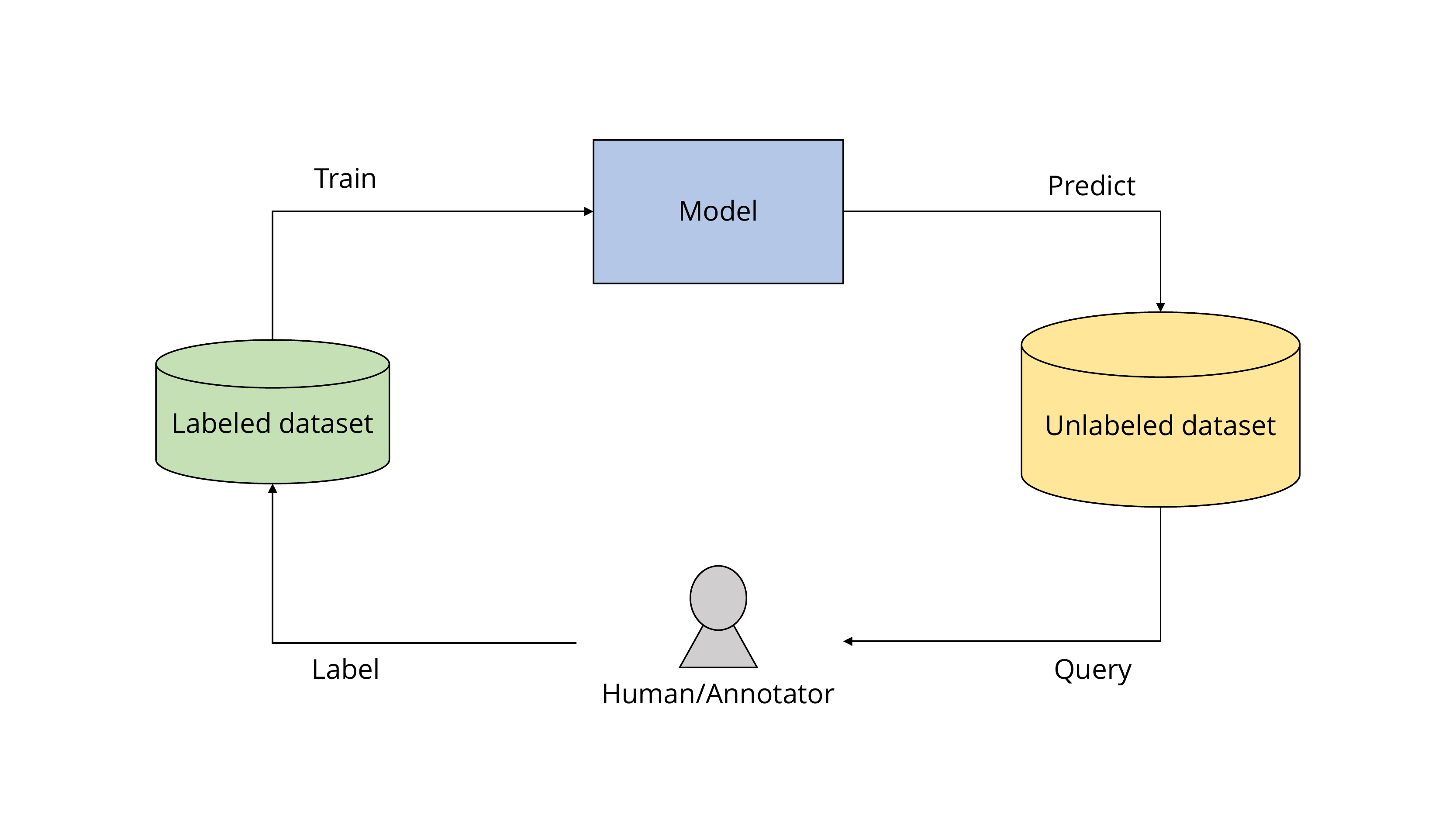}
	\caption{Pool-based active learning framework}\label{fig:al}
\end{figure}

\begin{itemize}

\item Uncertainty sampling (US) methods ---  Uncertainty sampling methods consider the model's prediction reliability of the data, and take the data with the largest uncertainty as the data for active learning query. \cite{lewis1994sequential,scheffer2001active,shannon2001mathematical}

\item Query by committee (QBC) methods ---  Query by committee methods are similar to ensemble methods in machine learning, voting on data based on the results predicted by multiple models and querying target data. \cite{seung1992query,dagan1995committee,mccallumzy1998employing}

\end{itemize}

\subsection{Incremental Learning}

The goal of incremental learning is to update the previous model to fit newly added samples and meanwhile not to lose the memory of the previous training. In other words, it is expected that the learning model can be updated to fit both the previous and newly added samples, when new samples are available. In the literature, there are three kinds of incremental learning strategies including the architecture-based manner \cite{mallya2018packnet}, the rehearsal-based manner \cite{rebuffi2017icarl, masana2020class} and the regularization-based manner \cite{kirkpatrick2017overcoming,li2017learning}. 

\begin{itemize}
\item The architecture-based manner equips new neurons into the previously-trained network when new samples are available, and then, freezing the other weights, use the new samples (sometimes with a part of previous samples) to train the weights associated with these neurons. When there are multi-round newly added samples, this manner will significantly enlarge the size of the network and influence the generalization performance of the final network.

\item The rehearsal-based manner selects a part of previous samples and combine them with the newly added samples to form new training samples. This manner could change the distribution of training samples and damage the previous training result of the learning model, which would not fit the previous samples yet.

\item The regularization-based manner introduces the regularization terms into the objective function to prevent the learning model from losing the memory of the previous training, when the previously-trained learning models are updated by using the new samples (sometimes, with a part of previous samples). 

\end{itemize}

Compared with the others, the regularization-based method has a high feasibility and is applicable to the incremental learning of many complicated deep learning models.

\subsection{Overview of Main Results}

\begin{figure}[htbp]
	\centering
	\includegraphics[width=4.5in]{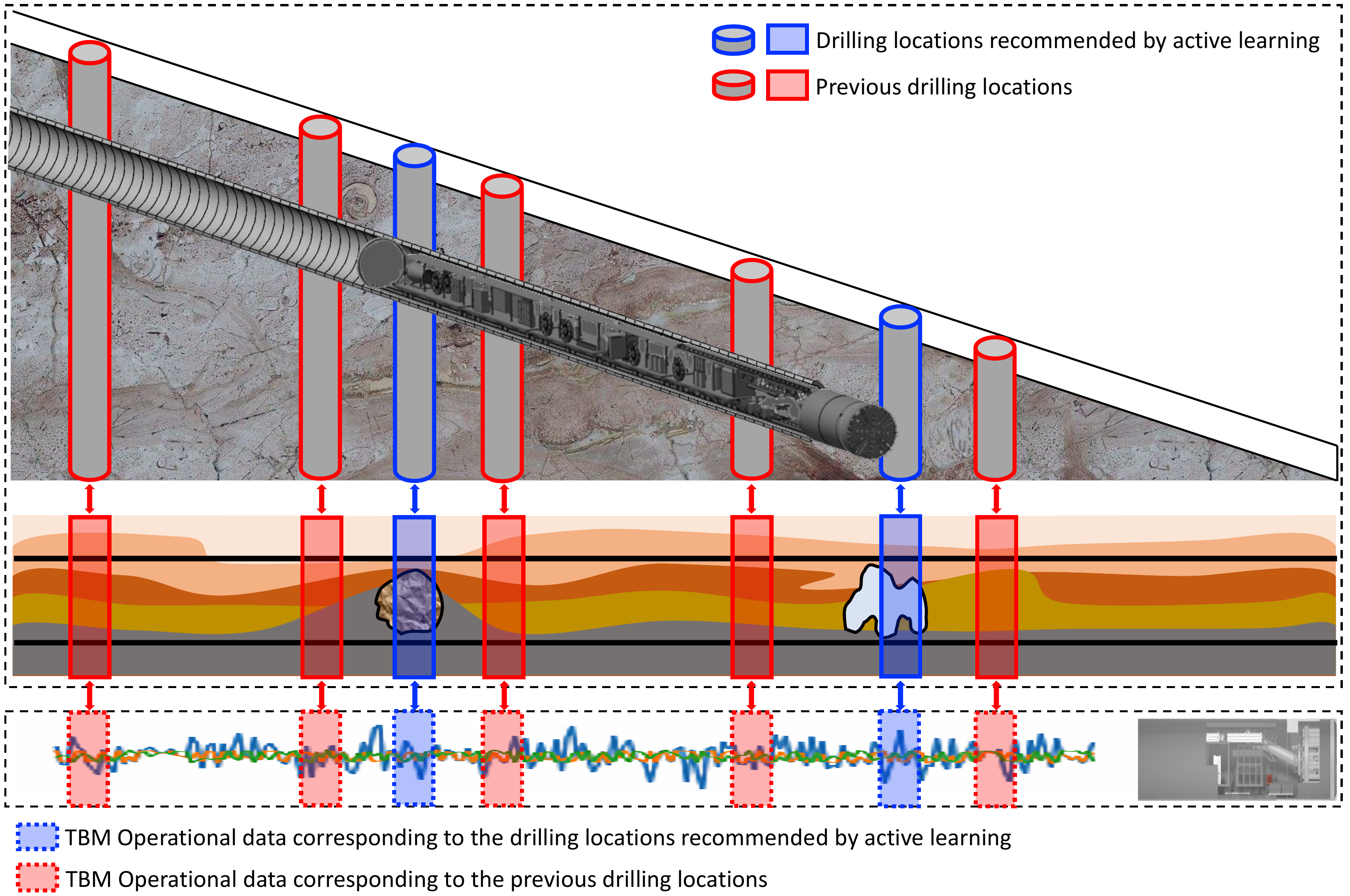}
	\caption{A diagram of active learning for TBM tunnel geological reconstruction. We first develop a reconstruction model by using the training samples composed of the geological information at the previous drilling locations (marked in red color) and the corresponding TBM operational data (marked in red color). If the resultant model performs unsatisfactorily, we further use the active learning method to recommend new drilling locations (marked in blue color) and then form new training samples, composed of the geological information at the previous drilling locations and the corresponding TBM operational data (marked in blue color), to update the previous model. }\label{fig:drill}
\end{figure}

As addressed above, the main challenge of tunnel geological reconstruction is the lack of labeled samples that can accurately describe the geological condition, especially the sudden changes, of the TBM construction tunnel. Taking advantages of the mechanism of actively acquiring informative samples, we propose an active learning-based framework, called AL-iGAN, for tunnel geological reconstruction to solve this issue ({\it cf.} Fig. \ref{fig:drill}). The proposed framework contains two parts: one is the usage of active learning methods and the other is a generative adversarial network for incremental learning.

The tunnel geological information is encoded into TBM operational data, collected from the sensors on TBM equipments during the construction phase and a small part of the operational data are labeled with the geological condition by using the drilling method. Unfortunately, the limited size of labeled TBM operational data cannot provide the satisfactory reconstruction results. The active learning methods are used to select the unlabeled operational data that will benefit to the geological reconstruction performance. The drilling method is used again to explore the geological condition at the tunnel locations associated with these unlabeled operational data. Subsequently, the selected operational data will be labeled for a new training phase. 

To improve the training efficiency and the reconstruction performance, we also design an incremental generative adversarial network for the geological reconstruction (iGAN-GR), which can be incrementally updated by using the new training samples rather than to be totally retrained. The proposed iGAN-GR is composed of two sub-networks: a generator and a discriminator. The former produces the estimate of the thickness of each rock-soil type appearing in the TBM construction tunnel, and the latter not only justifies whether the input data are the generator outputs or the real training samples but also checks whether the input data are fresh or previous training samples. Since TBM operational data have strong sequentiality and labeled samples are not sufficient, we adopt a positional-encoding attention layer \cite{gehring2017convolutional} in the generator of iGAN-GR to improve the reconstruction performance. In addition, following the regularization-based manner, we also adopt the elastic weight consolidation (EWC) loss, proposed in \cite{kirkpatrick2017overcoming}, as the regularization terms in the generative and the discriminative losses for training iGAN-GR. The numerical experiments support the effectiveness of the proposed framework and show that the active learning methods can effectively find the the most informative operational data that significantly improve the reconstruction performance.

The rest of this paper is organized as follows. In Section \ref{sec:al}, we summarize the query strategies of active learning used in this paper. In Section \ref{sec:gan}, we introduce the structure and training strategy of iGAN-GR. Section \ref{sec:main} presents the AL-iGAN framework for geological reconstruction. In Section \ref{sec:experiment}, we conduct the numerical experiments to examine the proposed framework and the last section concludes this paper.

\section{Incremental Generative Adversarial Network for Geological Reconstruction (iGAN-GR)}\label{sec:gan}

Generative adversarial networks (GANs), originally proposed in \cite{goodfellow2014generative}, are referred to a body of neural networks that are composed of two subnetworks: a generator and a discriminator, and trained by means of the minimax game between them. In general, the generator is a feedforward neural network (FNN) that implements the data (such as image and time series) generation tasks or the supervised learning tasks (such as classification and regression). The discriminator is also an FNN that roughly has a symmetric network structure of the generator. During the minimax game training, the discriminator will be trained to identify whether the output of generator is a real instance and the generator will be trained to provide a high-quality approximation of the real instance such that the discriminator is unable to identify the output of generator from the real instance. When reaching Nash equilibrium \cite{ratliff2013characterization}, the generator is deemed to fully capture the characteristics of the real-instance distribution. Taking advantages of the minimax game training where the discriminator guides the training of generator, many empirical evidences demonstrate that, given a sample set, the generator of GAN can perform better than an FNN with the same structure, trained by using the stochastic gradient descent (SGD) method \cite{liu2022gan, zhang2022scgan}.

Zhang {\it et al.} \cite{zhang2022generative} developed a GAN model, called GAN-GP, for geological prediction. However, this model is unsuitable (at least cannot be directly applied) to the geological reconstruction task considered in this paper. First, the labeled samples are fewer than those in the scenario of geological prediction; and second, there is no incremental learning mechanism in the training strategy of GAN-GP and thus it cannot be incrementally updated by using a limited amount of new labeled samples. To overcome these shortcomings, we designed an incremental generative adversarial network for geological reconstruction (iGAN-GR), whose weights can be incrementally updated by using the new samples recommended in the active learning phase. In Tab. \ref{tab:notation}, we summarize the main mathematical notations mentioned in this paper.

\begin{table}[htbp]
	\caption{List of mathematical notations}
	\label{tab:notation}
	\centering
	\resizebox{\linewidth}{!}{
		\begin{threeparttable} 
			\begin{tabular}{|c|c|c|c|}
				\hline
				Notation& Meaning & Notation& Meaning \\
				\hline
				$ d_{\bf{x}} $ &  Dimension of input $ {\bf x} $ & $ w_{0,p} $ & The $ p $-th weight of ${\rm net}_{0}$\\
				$ {\rm G}[\cdot] $ & iGAN-GR's generator   &$ \alpha_{0,p} $ & Change rate of gradient of $ w_p $ in $ {\rm net}_{0}(\cdot) $\\
				$ \textbf{y}_{\rm MSA} $ & Output of MSA module  &$ \mathcal{S}_{t} $ & Training dataset for $t$-th incremental training stage\\
				$ {\rm cnt}[\cdot] $ & Concatenation of vectors &$ {\rm net}_{t}^{k} $ & Network after the $ k $-th optimization iteration and the $ t $-th sample query\\
				 $ d_{\bf{y}}  $ &  Dimension of output $ {\bf y} $  &	$ \mathcal{L}_{\rm EWC} $ & Elastic weight consolidation loss\\
				$ L $ & Number of heads in MSA module&$ \mathcal{L}_{\rm IS} $ & Incremental supervised loss \\
				$ {\bf h}_l $ & The $ l $-th head in MSA module&$ {\cal L}^{(k)}_{{\rm G}}(\mathcal{S}_t) $ & Generative loss after the $k$-th iteration and $t$-th sampling query \\
				$ \sigma $ & Softmax function&$ {\cal L}^{(k)}_{{\rm D}}(\mathcal{S}_t) $ & discriminator loss after the $k$-th iteration and $t$-th sampling query \\
				$ \textbf{y}_{\rm PE} $ & Output of PE module&$ \eta_d $ & Learning rate for minimizing $ \mathcal{L}_{\rm D} $\\
				$ [\textbf{x}_{i}]_{2k} $ & The $ 2k $-th component of $ \textbf{x}_i $&$ \eta_g $ & Learning rate for minimizing $ \mathcal{L}_{\rm G} $\\
				$ \textbf{y}_{\rm FE} $ & Output of the feature extraction part&	$ \eta_s $ & Learning rate for minimizing $ \mathcal{L}_{\rm S} $\\
				$ I $ & the size of training samples &$ \lambda_d $ & Control factor corresponding to $ \mathcal{L}_{\rm EWC} $ for discriminator \\
				$ {\rm D}_{\rm r/g}[\cdot] $ & Discriminator output for real or generative instances&$ \lambda_g $ & Control factor corresponding to $ \mathcal{L}_{\rm EWC} $ for generator \\
				$ {\rm D}_{\rm o/f}[\cdot] $ & Discriminator output for original or fresh samples&$ \beta_d $ & Control factor corresponding to $ \mathcal{L}_{\rm IS} $ for discriminator \\
				$\mathcal{S}_0 = \left\{\textbf{x}_{(0,i)},{\bf y}_{(0,i)}\right\}_{i=1}^{I_0} $ & Original training samples &$ \beta_g $ & Control factor corresponding to $ \mathcal{L}_{\rm IS} $ for generator \\
				$ I_{t} $ & Training sample size for the $ t $-th incremental training &$ {\bf W}_{\rm G} $ & Generator weights\\
				$ \mathcal{L}_{\rm D} $ & Discriminative loss&	$ {\bf W}_{\rm D} $ & Discriminator weights\\
				$ \mathcal{L}_{\rm G} $ & Generative loss&	$ \textbf{x}_{(t,j)}^{\rm add} $ & the $ j $-th point in $ \mathcal{S}_t^{add} $\\
				$ \mathcal{L}_{\rm S} $ & Supervised loss&$ {\rm D}^{(k)} $ & Discriminator after the $ k $-th optimization iteration\\
				$ \left\|\cdot \right\|_{2} $ & Euclidean norm&$ {\rm G}^{(k)} $ & Generator after the $ k $-th optimization iteration\\
				$ \mathcal{S}_t^{add}= \left\{\textbf{x}_{(t,j)}^{\rm add},{\bf y}_{(t,j)}^{\rm add}\right\}_{j=1}^{J_t}$ & Newly-added training dataset in the $ t $-th sample query&$ \mathcal{L}_{\rm D}^{(k)} $ & Discriminative loss after the $ k $-th optimization iteration\\
				$ \tau[\cdot] $ & Indicator function for original or fresh samples &$ \mathcal{L}_{\rm G}^{(k)} $ & Generative loss after the $ k $-th optimization iteration\\
				$ {\rm net}_{0}(\cdot) $ & Pre-existing network for initial training stage &$ \textbf{x}'_{\rm EN} $ & Point queried via entropy sampling\\
				$ \mathcal{P} = \{{\bf x}^{\rm pool}_m\}$ & Data pool&$ \textbf{x}'_{\rm VE} $ &Point queried via vote entropy based on QBC\\
				$ \textbf{{x}}_{(t,i)} $  & the $ i $-th point in the $ t $-th sample query  &$ \textbf{x}_{t}^{\rm query}$& Point selected in the $ t $-th sample query\\
				$ \mathcal{C} $ & Committee in QBC method &iGAN-GR$_{t} $ & Network after $ t $-th sampling query\\
				\hline
			\end{tabular}
		\end{threeparttable}
	}
\end{table}

\subsection{Structure of iGAN-GR}

\begin{figure}[htbp]
	\centering
	\includegraphics[width=4.5in]{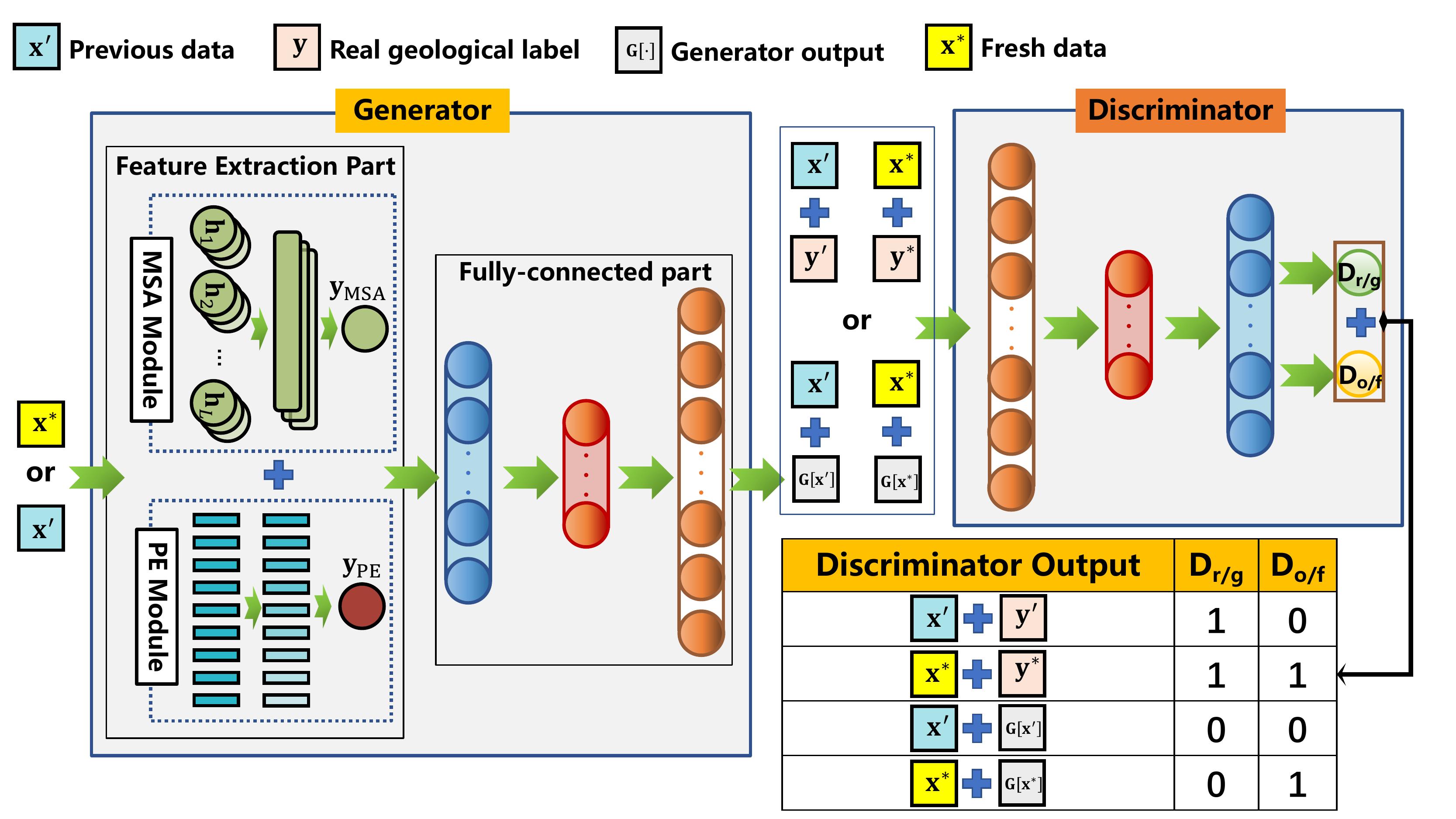}
	\caption{Structure of iGAN-GR}\label{fig:gan}
\end{figure}
Similar to the structure of traditional GANs, iGAN-GR is also composed of two sub-networks: a generator and a discriminator ({\it cf.} Fig. \ref{fig:gan}). Next, we will demonstrate the structure characteristics of iGAN-GR's generator and discriminator, respectively.

The generator of iGAN-GR, denoted as ${\rm G}[\cdot]$, consists of two parts: the feature extraction part and the regression part. The former is a combination of the multi-head self-attention (MSA) module and the position encoding (PE) module in parallel. Different for the ordinary convolution operation that detects the local features from the patches of the input data, the MSA module captures the global features of the input data by mapping them into different subspaces spanned by the weight vectors. Specifically, given an input ${\bf x}\in\mathbb{R}^{d_{\bf x}}$, the output of the MSA is expressed as
\begin{align}\label{eq:msa}
	{\bf y}_{\rm MSA}(\textbf{x}) = & {\bf W}^{O}\cdot {\rm cnt}[{\bf h}_1,\cdots,{\bf h}_L]; \nonumber \\
	{\bf h}_l = & \sigma\left(\frac{\textbf{x}^T ({\bf W}_l^{Q})^T{\bf W}_l^{K}\textbf{x}}{\sqrt {d_{\bf x}} }\right)\cdot {\bf W}_l^{V} \textbf{x} ,\;\; l=1,\cdots,L,
\end{align}
where $\sigma:\mathbb{R} \rightarrow \mathbb{R}$ is the softmax function; ${\rm cnt}[\cdot] $ stands for the concatenation of vectors; $L$ is the number of heads; and ${\bf W}_{l}^{Q},{\bf W}_{l}^{K}, {\bf W}_{l}^{V},{\bf W}^{O} $ are undetermined weights. There is a caveat that the individual usage of MSA module is possible to miss some important local features of the data, especially for the data that have strong sequentiality. Therefore, the PE module, which detects the positional relationship among the different inputs, is employed to collaborate with the MSA module:
\begin{equation}
	\textbf{y}_{\rm PE}(\textbf{x}_i) = {\rm cnt}[{\rm PE}([{\textbf{x}_i]_1}),\cdots,{\rm PE}([{\textbf{x}_i]_{d_{\bf x}}})]
\end{equation}
with
\begin{equation}
	\left\{\begin{matrix}{\rm PE}([\textbf{x}_i]_{2k}) = \sin(\frac{i}{10000^{2k/{d_{\bf x}}}})
		\\{\rm PE}([\textbf{x}_i]_{2k+1}) = \cos(\frac{i}{10000^{2k/{d_{\bf x}}}})
	\end{matrix},\right.\nonumber
\end{equation}
where $[\textbf{x}_i]_{2k}$ (resp. $[\textbf{x}_i]_{2k+1} $) is the $2k$-th (resp. $2k+1$-th) component of the vector $\textbf{x}_i\in\mathbb{R}^{d_{\bf x}}$ and $i\in \{1,\cdots, I\}$ is the index of the input ${\bf x}_i$. It is noteworthy that the structure of PE module is fixed and there is no undetermined weight to be trained during the training phase. Moreover, given an input ${\bf x}_i$ ($i\in \{1,2,\cdots, I\}$), the output of the feature extraction part is expressed as: 
\begin{equation*}
{\bf y}_{\rm FE}({\bf x}_i) = {\bf y}_{\rm MSA}(\textbf{x}) + \textbf{y}_{\rm PE}(\textbf{x}_i).
\end{equation*}
Then, the regression part is a stack of several fully-connected layers and each node of its output layer signifies the relative thickness of each rock-soil type appearing in the TBM construction tunnel. 

The discriminator of iGAN-GR is also a stack of fully-connected layers and usually roughly symmetric with the structure of the regression part of the generator. The input of the discriminator is either a real instance ({\it i.e.,} the pair of an input vector and its real geological labels) or a generative instance ({\it i.e.,} the pair of an input vector and the corresponding generator output). Different from the traditional GANs, the discriminator of iGAN-GR has two output nodes of $0$-$1$ form, denoted as ${\rm D}_{\rm r / g}[\cdot]$ and ${\rm D}_{\rm o/f}[\cdot]$, respectively. The first one ${\rm D}_{\rm r/g}[\cdot]$ indicates whether the input pair is a real instance, labeled with ``1" or a generative instance, labeled with ``0"; and the second one ${\rm D}_{\rm o/f}[\cdot]$ indicates whether the input pair is derived from the fresh training data that are recommended by the active learning models, labeled with ``1" or from the original or previous training data, labeled with ``0".

Subsequently, we will address the training strategy of iGAN-GR that contains two stages: the initial training stage and the incremental training stage. The former is to initially train iGAN-GR based on the existed samples; and the latter is to incrementally update the weights of iGAN-GR by using the fresh training data that are recommended by the active learning models.

\subsection{Initial Training Stage of iGAN-GR}

Let $ \mathcal{S}_0:=\left\{\textbf{x}_{(0,i)},{\bf y}_{(0,i)}\right\}_{i=1}^{I_0} \subset \mathbb{R}^{d_\textbf{x}}\times\mathbb{R}^{d_\textbf{y}} $ be the original training dataset. In the initial training stage, since there are no newly-added training data, we only need to consider the first discriminator output ${\rm D}_{\rm r/g}[\cdot]$. Then, the discriminative loss is designed as follows:
\begin{equation}
	{\cal L}_{{\rm D}}(\mathcal{S}_0) = \frac{1}{I_0}\sum_{i=1}^{I_0} \log (1-{\rm D}_{\rm r/g} [\textbf{x}_{(0,i)},{\bf y}_{(0,i)}]) + \log ({\rm D}_{\rm r/g} [\textbf{x}_{(0,i)},{\rm G}[\textbf{x}_{(0,i)}]).
\end{equation}
The minimization of $ {\cal L}_{{\rm D}}(\mathcal{S}_0) $ aims to improve the reconstruction performance of generator and meanwhile to make the discriminator more accurately identify whether its input is a real instance.

On the other hand, the generative loss is defined as:
\begin{equation}
	{\cal L}_{{\rm G}}(\mathcal{S}_0) = \frac{1}{I_0}\sum_{i=1}^{I_0} \log (1-{\rm D}_{\rm r/g}[\textbf{x}_{(0,i)},{\rm G}[\textbf{x}_{(0,i)}]]).
\end{equation}
The minimization of $ {\cal L}_{{\rm G}}(\mathcal{S}_0) $ will make the generator accurately approximate the distribution of training data so as to fake the discriminator.

To enhance the training stability, we also introduce the supervised loss ${\cal L}_{{\rm S}}(\mathcal{S}_0)$ to balance the training levels of the generator and the discriminator in each iteration:
\begin{equation}
	{\cal L}_{{\rm S}}(\mathcal{S}_0) =  \frac{1}{I_0}\sum_{i=1}^{I_0} \left\|{\rm G}[\textbf{x}_{(0,i)}]-{\bf y}_{(0,i)} \right\|_{2}^{2},
\end{equation}
where $  \left\|\cdot \right\|_{2} $ stands for the Euclidean norm. During each iteration of the adversarial training process, we minimize the supervised loss $ {\cal L}_{{\rm S}} $ to refine the generator weights to prevent the generator to be misguided by the discriminator that is not well trained. These loss functions will be minimized by using minibatch Adam method \cite{kingma2014adam} and the main workflow of iGAN-GR's initial training stage is illustrated in Alg. \ref{alg:initial}.

\begin{algorithm}[htbp]
	\renewcommand{\algorithmicrequire}{\textbf{Input:}}
	\renewcommand{\algorithmicensure}{\textbf{Output:}}
	\caption{Workflow of Initially Training iGAN-GR}\label{alg:initial}
	\begin{algorithmic}[1]	
		\REQUIRE $\mathcal{S}_0$, $K$, $\eta_g$, $\eta_d$, $\eta_s$; 
		\ENSURE generator weights ${\bf W}_{\rm G}$, discriminator weights ${\bf W}_{\rm D}$;
		\STATE Randomly initialize ${\bf W}_{\rm G}$ and ${\bf W}_{\rm D}$; 
		\STATE Minimize the supervised loss ${\cal L}_{{\rm S}}(\mathcal{S})$ to pre-train the generator and obtain the generator weights ${\bf W}^{(0)}_{\rm G}$;
		\FORALL{ $k=1,2,\cdots,K$; }
		\STATE Update ${\bf W}^{(k-1)}_{\rm G}$ by minimizing the supervised loss ${\cal L }_{{\rm S}}(\mathcal{S}_0)$: ${\bf W}_{\rm G}^{(k,1)} = {\bf W}_{\rm G}^{(k-1)} - \eta_s\cdot\frac{\partial  {\cal L}_{\rm S}(\mathcal{S}_0)}{\partial {\bf W}_{\rm G}^{(k-1)}};$
		\STATE Update ${\bf W}_{\rm G}^{(k,1)}$ and ${\bf W}_{\rm D}$ by minimizing the discriminative loss ${\cal L}_{\rm D}(\mathcal{S}_0)$: ${\bf W}_{\rm G}^{(k,2)} = {\bf W}_{\rm G}^{(k,1)} -  \eta_d\cdot\frac{\partial  {\cal L}_{\rm D}(\mathcal{S}_0)}{\partial {\bf W}_{\rm G}^{(k,1)}}$ and ${\bf W}_{\rm D}^{(k,1)} = {\bf W}_{\rm D}^{(k)} -  \eta_d\cdot\frac{\partial  {\cal L}_{\rm D}(\mathcal{S}_0)}{\partial {\bf W}_{\rm D}}$; 
		\STATE Update  ${\bf W}_{\rm G}^{(k,2)}$ by minimizing the supervised loss ${\cal L}_{\rm S}(\mathcal{S}_0)$: ${\bf W}_{\rm G}^{(k,3)} = {\bf W}_{\rm G}^{(k,2)} -  \eta_s\cdot\frac{\partial  {\cal L}_{\rm S}(\mathcal{S}_0)}{\partial {\bf W}_{\rm G}^{(k,2)}};$ 
		\STATE Update ${\bf W}_{\rm G}^{(k,3)}$ and ${\bf W}_{\rm D}^{(k,1)}$ by minimizing the generative loss ${\cal L}_{\rm G}(\mathcal{S}_0)$: ${\bf W}_{\rm G}^{(k,4)} = {\bf W}_{\rm G}^{(k,3)} -  \eta_g\cdot\frac{\partial  {\cal L}_{\rm G}(\mathcal{S}_0)}{\partial {\bf W}_{\rm G}^{(k,3)}}$ and ${\bf W}_{\rm D}^{(k+1)} = {\bf W}_{\rm D}^{(k,1)} -  \eta_g\cdot\frac{\partial  {\cal L}_{\rm G}(\mathcal{S}_0)}{\partial {\bf W}_{\rm D}^{(k,1)}}$; 
		\STATE Update  ${\bf W}_{\rm G}^{(k,4)}$ by minimizing the supervised loss ${\cal L}_{\rm S}(\mathcal{S}_0)$: ${\bf W}_{\rm G}^{(k+1)} = {\bf W}_{\rm G}^{(k,4)} - \eta_s\cdot \frac{\partial  {\cal L}_{\rm S}(\mathcal{S}_0)}{\partial {\bf W}_{\rm G}^{(k,4)}};$
		\ENDFOR
	\end{algorithmic}
\end{algorithm}

\subsection{Incremental Training Stage of iGAN-GR}

Let $ \mathcal{S}_t^{\rm add}:=\left\{\textbf{x}^{\rm add}_{(t,j)},\textbf{y}^{\rm add}_{(t,j)}\right\}_{j=1}^{J_t} \subset \mathbb{R}^{d_\textbf{x}}\times\mathbb{R}^{d_\textbf{y}} $ be the newly-added training dataset that are recommended by using the active learning model for the $t$-th time. Define an indicator function $\tau[\cdot] : \mathbb{R}^{d_\textbf{x}}\times\mathbb{R}^{d_\textbf{y}}  \rightarrow \{0,1\}$ as
\begin{equation*}
\tau[(\textbf{x},{\bf y})] = 
\left\{
\begin{array}{ll}
  1,&  \mbox{when $(\textbf{x},{\bf y}) \in  \mathcal{S}_t^{\rm add}$ ($\forall \, t \geq 1$)};   \\
  0, &  \mbox{when $(\textbf{x},{\bf y}) \in  \mathcal{S}_0$},
\end{array}
\right.
\end{equation*}
which signifies whether a training datum $(\textbf{x},{\bf y})$ is newly-added.

In the incremental training stage, we adopt the elastic weight consolidation (EWC), a regularization-based incremental learning method proposed in \cite{kirkpatrick2017overcoming}, to update the iGAN-GR's weights
when the newly-added training data are available. Let ${\rm net}_0(\cdot):\mathbb{R}^{d_{\bf x}} \rightarrow \mathbb{R}^{d_{\bf y}}$ be a pre-existing network with $P$ weights $w_{0,1},\cdots,w_{0,P}$ that has been trained by using the original training dataset $\mathcal{S}_0:=\left\{\textbf{x}_{(0,i)},{\bf y}_{(0,i)}\right\}_{i=1}^{I_0}$, and denote 
\begin{equation*}
\alpha_{0,p} = \frac{1}{I_0}\sum_{i=1}^{I_0} \frac{\partial^2  \, \| {\rm net}_0(\textbf{x}_{(0,i)})  -  {\bf y}_{(0,i)}  \|_2^2  }{\partial^2 \,w_{0,p }},
\end{equation*}
which recodes the change rate of gradient of the $p$-th weight $w_p$ in the previous process of training ${\rm net}_0(\cdot)$. A large $\alpha_{0,p}$ means that the weight $w_{0,p}$ will be significantly updated in the training process. Then, combining a part (usually $80\%$) of the original training dataset $\mathcal{S}_0$ with the newly-added dataset $\mathcal{S}_1^{\rm add}$ to form a new training dataset $\mathcal{S}_1:=\left\{\textbf{x}_{(1,i)},{\bf y}_{(1,i)}\right\}_{i=1}^{I_1} \subset \mathcal{S}_0  \cup \mathcal{S}_1^{\rm add}$, we implement a new process of training ${\rm net}_0(\cdot)$. Let ${\rm net}_1^{(k)}(\cdot)$ be the network after the $k$-th iteration and $w^{(k)}_{1,1},\cdots,w^{(k)}_{1,P}$ be the newly-updated weights accordingly. The EWC loss of the newly-updated network is expressed as
\begin{equation*}
	{\cal L}_{\rm EWC}({\rm net}_1^{(k)}) = \frac{1}{2} \sum_{p=1}^{P} |\alpha_{0,p}|  \cdot (w^{(k)}_{1,p}-w_{0,p})^2.
\end{equation*}
The minimization of ${\cal L}_{\rm EWC}({\rm net}_1^{(k)})$ aims to control the updating amplitude of the weights that sensitive in the training process. In this manner, the previous weights $w_{0,p}$ are used to control the training behavior of the newly-updated network ${\rm net}_1^{(k)}$ and then to prevent the new network ${\rm net}_1^{(k)}$ from forgetting the previous training information.

During the new training process, the discriminator is also equipped with an extra output node ${\rm D}_{\rm o/f}[\cdot]$, which indicates  whether the input pair is derived from the fresh training dataset $\mathcal{S}_1^{\rm add}$ or from the previous training dataset $\mathcal{S}_0$. Accordingly, we introduce the incremental supervised (IS) loss:
\begin{align*}
	{\cal L}_{\rm IS}({\cal S}_1) = & \frac{1}{2I_1}\sum_{i=1}^{I_1} \Big( ({\rm D}_{\rm o/f}[\textbf{x}_{1,i},{\rm G}[\textbf{x}_{1,i}]]-\tau[(\textbf{x}_{1,i},{\bf y}_{1,i})] )^{2} \nonumber\\
	&\qquad\qquad\qquad\qquad+ ({\rm D}_{\rm o/f}[\textbf{x}_{1,i},{\bf y}_{1,i}]-\tau[(\textbf{x}_{1,i},{\bf y}_{1,i})] )^{2} \Big).
\end{align*}
The minimization of $ {\cal L}_{\rm IS}({\cal S}) $ aims to make the discriminator distinguish between the fresh data and the previous data and meanwhile the generator can be guided not only to fit the fresh training dataset $\mathcal{S}_1^{\rm add}$ but also to maintain the memory on the previous training dataset $\mathcal{S}_0$.

Overall, in the incremental training stage, the generative loss after the $k$-th iteration is expressed as
\begin{equation}
	{\cal L}^{(k)}_{{\rm G}}(\mathcal{S}_1) =  {\cal L}_{{\rm G}}(\mathcal{S}_1) + \lambda_g {\cal L}_{{\rm EWC}}({\rm G}^{(k)}) + \beta_g {\cal L}_{{\rm IS}}(\mathcal{S}_1),
\end{equation}
where ${\rm G}^{(k)}$ stands for the generator after the $k$-th iteration; and $\lambda_g>0$ (resp. $\beta_g>0$) is the control factor corresponding to the EWC (resp. IS) loss. Similarly, the discriminative loss for incremental training is given by
\begin{equation}
	{\cal L}^{(k)}_{{\rm D}}(\mathcal{S}_1) =  {\cal L}_{{\rm D}}(\mathcal{S}_1) + \lambda_d {\cal L}_{{\rm EWC}}({\rm D}^{(k)})  + \beta_d {\cal L}_{{\rm IS}}(\mathcal{S}_1),
\end{equation}
where ${\rm D}^{(k)}$ stands for the discriminator after the $k$-th iteration; and $\lambda_d>0$ (resp. $\beta_d>0$) is the control factor corresponding to the EWC (resp. IS) loss. In addition, the supervised loss $ \mathcal{L}_S(\mathcal{S}) $ is also employed to balance the training levels between the generator and the discriminator in each iteration during the incremental learning stage. Both of the two loss functions ${\cal L}^{(k)}_{{\rm G}}(\mathcal{S})$ and ${\cal L}^{(k)}_{{\rm D}}(\mathcal{S})$ are also optimized by using minibatch Adam method and the workflow is illustrated in Alg. \ref{alg:incremental}

\begin{algorithm}[htbp]
	\renewcommand{\algorithmicrequire}{\textbf{Input:}}
	\renewcommand{\algorithmicensure}{\textbf{Output:}}
	\caption{Workflow of Incrementally Training iGAN-GR}\label{alg:incremental}
	\begin{algorithmic}[1]	
		\REQUIRE $\mathcal{S}$, $K$,  $ \lambda_g $, $  \lambda_d $, $ \beta_g $, $ \beta_d $,  $\eta_g$, $\eta_d$, $\eta_s$, ${\bf W}^{(0)}_{\rm G}$, ${\bf W}^{(0)}_{\rm D}$; 
		\ENSURE generator weights ${\bf W}_{\rm G}$, discriminator weights ${\bf W}_{\rm D}$;
		\FORALL{ $k=1,2,\cdots,K$; }
		\STATE Update ${\bf W}^{(k-1)}_{\rm G}$ by minimizing the supervised loss ${\cal L}_{{\rm S}}(\mathcal{S})$: ${\bf W}_{\rm G}^{(k,1)} = {\bf W}^{(k-1)}_{\rm G} - \eta_s\cdot\frac{\partial  {\cal L}_{\rm S}(\mathcal{S})}{\partial {\bf W}^{(k-1)}_{\rm G}};$
		\STATE Update ${\bf W}_{\rm G}^{(k,1)}$ and ${\bf W}_{\rm D}^{(0)}$ by minimizing the discriminative loss ${\cal L}^{(k)}_{{\rm D}}(\mathcal{S})$: ${\bf W}_{\rm G}^{(k,2)} = {\bf W}_{\rm G}^{(k,1)} -  \eta_d\cdot\frac{\partial  {\cal L}_{\rm D}^{(k)}(\mathcal{S})}{\partial {\bf W}_{\rm G}^{(k,1)}}$ and ${\bf W}_{\rm D}^{(k,1)} = {\bf W}_{\rm D}^{(k)} -  \eta_d\cdot\frac{\partial  {\cal L}_{\rm D}^{(k)}(\mathcal{S})}{\partial {\bf W}_{\rm D}^{(k)}}$; 
		\STATE Update  ${\bf W}_{\rm G}^{(k,2)}$ by minimizing the supervised loss ${\cal L}_{\rm S}(\mathcal{S})$: ${\bf W}_{\rm G}^{(k,3)} = {\bf W}_{\rm G}^{(k,2)} -  \eta_s\cdot\frac{\partial  {\cal L}_{\rm S}(\mathcal{S})}{\partial {\bf W}_{\rm G}^{(k,2)}};$ 
		\STATE Update ${\bf W}_{\rm G}^{(k,3)}$ and ${\bf W}_{\rm D}^{(k,1)}$ by minimizing the generative loss ${\cal L}^{(k)}_{{\rm G}}(\mathcal{S})$: ${\bf W}_{\rm G}^{(k,4)} = {\bf W}_{\rm G}^{(k,3)} -  \eta_g\cdot\frac{\partial  {\cal L}_{\rm G}^{(k)}(\mathcal{S})}{\partial {\bf W}_{\rm G}^{(k,3)}}$ and ${\bf W}_{\rm D}^{(k+1)} = {\bf W}_{\rm D}^{(k,1)} -  \eta_g\cdot\frac{\partial  {\cal L}_{\rm G}^{(k)}(\mathcal{S})}{\partial {\bf W}_{\rm D}^{(k,1)}}$; 
		\STATE Update  ${\bf W}_{\rm G}^{(k,4)}$ by minimizing the supervised loss ${\cal L}_{\rm S}(\mathcal{S})$: ${\bf W}_{\rm G}^{(k+1)} = {\bf W}_{\rm G}^{(k,4)} - \eta_s\cdot \frac{\partial  {\cal L}_{\rm S}(\mathcal{S})}{\partial {\bf W}_{\rm G}^{(k,4)}};$
		\ENDFOR
	\end{algorithmic}
\end{algorithm}

\section{AL-iGAN Framework for Tunnel Geological Reconstruction}\label{sec:main}

In this section, we show the workflow of the proposed AL-iGAN framework for tunnel geological reconstruction. First, we introduce some preliminaries on active learning.

\subsection{Query Strategies of Active Learning}\label{sec:al}

Here, we adopt the active learning with pool-based sampling, where the fresh points will be selected from a pool $\mathcal{P} = \{{\bf x}^{\rm pool}_m\}$ of inputs of candidate samples based on a query strategy, and then a human annotator will label these points to form new training samples. The query strategy is the key to active learning, and there are two kinds of the most frequently used query strategies: one is uncertainty sampling, which selects the points making the outputs of the current learning model have the greatest degree of uncertainty \cite{lewis1994sequential}; and the other is query-by-committee (QBC), which selects the points according to the voting results of the committee members \cite{seung1992query}.

\subsubsection{Entropy-based Uncertainty-sampling (EUS) Strategy}

Given a pool $ \mathcal{P} = \{{\bf x}^{\rm pool}_m\}$, let $\mathcal{M}_\mathcal{S}(\cdot): \mathbb{R}^{d_x} \rightarrow  \mathbb{R}^{d_y}$ stand for the current learning model that has been trained by using the original samples $\mathcal{S}_0:=\left\{\textbf{x}_{(0,i)},{\bf y}_{(0,i)}\right\}_{i=1}^{I_0}$. Uncertainty sampling selects a point ${\bf x}' \in \mathcal{P}$ such that the predicted value $\left[\mathcal{M}_\mathcal{S}(\textbf{x}') \right]_j (j = \{1,2,\cdots,d_y\})$ have the smallest discrepancies, {\it i.e.,} the learning model $\mathcal{M}_\mathcal{S}$ has the largest uncertainty at the point ${\bf x}'$. Entropy sampling (ES) \cite{scheffer2001active} chooses the point ${\bf x}_{\rm EN}'$ from the pool $\mathcal{P}$ in the way of
\begin{equation}\label{eq:EN}
	\textbf{x}'_{\rm EN } := \mathop{\arg\max}\limits_{\textbf{x}'\in\mathcal{P}}\left\{ -\sum_{j=1}^{d_y} \left[\mathcal{M}_\mathcal{S}(\textbf{x}') \right]_j \log \left[\mathcal{M}_\mathcal{S}(\textbf{x}') \right]_j \right\}.
\end{equation}
Namely, the resultant point $\textbf{x}'_{\rm EN}$ should make the distribution of predicted value $\left[\mathcal{M}_\mathcal{S}(\textbf{x}') \right]_j$ is closer to the uniform distribution than the other points in the pool.

\subsubsection{Query-by-Committee (QBC) Strategy}
One main shortcoming of uncertainty-sampling strategy is the computation of predicted value, which usually brings a heavy computational burden. Instead, QBC strategy sets up a committee $ \mathcal{C} =\{ \mathcal{M}^1,\cdots,\mathcal{M}^C \}$ and the committee members $\mathcal{M}^c$ are auxiliary learning models trained by using the original training samples. Each committee member votes on the rock-soil type of the maximum predicted value of the candidate input points in the pool ${\cal P}$, and the input point with the highest level of disagreement among these voting results will be recommended to form a new training sample. Based on the vote entropy (VE) \cite{dagan1995committee}, the point ${\bf x}_{\rm VE}'$ is selected from the pool $\mathcal{P}$ in the way of
\begin{equation}\label{eq:VE}
	\textbf{x}'_{\rm VE } = \mathop{\arg\max}_{\textbf{x}\in\mathcal{P}} \left\{ -\sum_{j=1}^{d_y} \frac{V(j)}{C}\log \frac{V(j)}{C} \right\},
\end{equation}
where $V(j)$ stands for the number of voting results that are the $j$-th label.
\begin{remark}
We normalized the thicknesses of each rock-soil types, and QBC is designed to be more concerned about the data with multiple rock-soil types which are harder to fit and predict than just one.
\end{remark}

\subsection{Workflow of AL-iGAN Framework}

Given an original training dataset ${\cal S}_0 := \{ {\bf x}_{(0,i)}, {\bf y}_{(0,i)}\} _ {i=1}^{I_0}$, we first train the iGAN-GR model via the workflow given in Alg. \ref{alg:initial} and the resultant network is denoted as {\rm iGAN-GR}$_0$. Then, we adopt the active learning method to select the most informative operational datum ${\bf x}_1^{\rm query}$ from the operational data pool ${\cal P}$ and then to retrieve the location corresponding to the selected data ${\bf x}_1^{\rm query}$ in the construction tunnel. The operational data within $0.3$m around this drilling location are labeled with the present rock-soil types and then gather them to form the newly-added training dataset ${\cal S}^{\rm add}_1 =\{ {\bf x}_{(1,j)}, {\bf y}_{(1,j)}\} _ {j=1}^{J_1}$. Next, we randomly take $80\%$ from the original training dataset $\mathcal{S}_0$ and then combine them with the set ${\cal S}^{\rm add}_1 $ to form a new training dataset ${\cal S}_1$. We use it to incrementally train the network iGAN-GR$_0$ via the workflow given in Alg. \ref{alg:incremental}. The resultant network is denoted as {\rm iGAN-GR}$_1$ and is further used to query anther operational datum ${\bf x}_2^{\rm query}$ from the pool ${\cal P} \setminus \{  {\bf x}_{(1,j)} \}_{j=1}^{J_1}$. The combination of a $80\%$ part of ${\cal S}_1$ and the set ${\cal S}^{\rm add}_2$ derived from ${\bf x}_2^{\rm query}$ leads to an other new training dataset ${\cal S}_2$, which will be used to incrementally train the network iGAN-GR$_2$. 

Repeatedly, denote ${\bf x}_t^{\rm query}$ as the operational datum selected in the $t$-th sample querying of the active learning method and $ \mathcal{S}_t^{\rm add}:=\left\{\textbf{x}^{\rm add}_{(t,j)},\textbf{y}^{\rm add}_{(t,j)}\right\}_{j=1}^{J_t} \subset \mathbb{R}^{d_\textbf{x}}\times\mathbb{R}^{d_\textbf{y}}  $ as the corresponding newly-added training dataset. Let {\rm iGAN-GR}$_t$ stands for the network after the $t$-th incremental training based on the training dataset ${\cal S}_t $ composed of a $80\%$ part of ${\cal S}_{t-1}$ and the set ${\cal S}^{\rm add}_t$. We repeat the aforementioned process of incrementally training based on the active learning method until the terminal condition is satisfied. The workflow of AL-iGAN framework is illustrated in Alg. \ref{alg:al-igan}. 

\begin{algorithm}[htbp]
	\renewcommand{\algorithmicrequire}{\textbf{Input:}}
	\renewcommand{\algorithmicensure}{\textbf{Output:}}
	\caption{Workflow of AL-iGAN framework}\label{alg:al-igan}
	\begin{algorithmic}[1]	
		\REQUIRE $\mathcal{S}_0$, {\rm iGAN-GR}$_0$, $ {\cal P} $; 
		\ENSURE {\rm iGAN-GR} model;
		\FORALL{ $t=1,2,\cdots,T$; }
		\STATE Used {\rm iGAN-GR}$_{t-1} $to query the most informative operational datum $ {\bf{x}}_{t}^{\rm query} $ from $ {\cal P}  \setminus \{ \{  {\bf x}_{(1,j)} \}_{j=1}^{J_1}\cup \cdots \cup \{{\bf x}_{(t,j)} \}_{j=1}^{J_t}\}$.
		\STATE Gather the operational data within $0.3$m around this drilling location to form the newly-added training dataset ${\cal S}^{\rm add}_t$.
		\STATE Randomly take $80\%$ from $\mathcal{S}_{t-1}$ and then combine them with the set ${\cal S}^{\rm add}_t $ to form ${\cal S}_t$.
		\STATE Use ${\cal S}_t$ to incrementally train the network iGAN-GR$_{t-1}$ via the workflow given in Alg. \ref{alg:incremental} and get {\rm iGAN-GR}$_t$. 
		\ENDFOR
	\end{algorithmic}
\end{algorithm}

\section{Numerical Experiments}\label{sec:experiment}

We conduct the numerical experiments to validate the proposed AL-iGAN framework for geological construction based on a real-world TBM operational data that are partially labeled with geological information. In the experiments, we mainly concern with the following issues:
\begin{itemize}
	\item whether iGAN-GR performs better than the state-of-the-art learning models in the geological reconstruction tasks;
	\item whether the active learning can effectively improve the reconstruction performance;
	\item the comparison between different query strategies of active learning. 
	\item the effect of incremental learning in the geological reconstruction tasks.

\end{itemize}
All experiments are performed on a computer equipped with Intel\circled{R}i9-10850K CPU at 3.60 GHz×8, 64GB RAM and two Nvidia\circled{R}GTX-3080 graphic cards.

\subsection{Data Preparation}

The operational data are collected by sensors on the earth pressure balance shield machine that was operated in an urban subway construction project. The tunnel is about $2000$m long with the diameter of $6.3$ meters and the project route is divided into $1364$ ring sections, each of which is about $1.5$m long. The ground elevation is $0.14\sim5.86$m and the tunnel ground depth is within $11.8\sim25.4$m. There are a total of $4.6$ million operational data, and each data has $d_\textbf{x}=69$ attributes corresponding to different operational parameters, such as torque, thrust, tunneling speed and fuel tank temperature.

The geological information is detected by using the drilling method at $88$ locations at the ground surface above the construction tunnel. There are $11$ rock-soil types appearing in the geological samples. They are signified by using different values of $7$ physical and mechanical indicators given in Tab. \ref{tab:rock-soil}. The thickness of the $d_{\bf y} = 11$ rock-soil types is set as the geological reconstruction result provided by iGAN-GR's generator.

The operational data within $0.3$m around a drilling location are labeled with the rock-soil types appearing at the drilling point. We select $60$ out of the $88$ drilling locations to form the training and testing dataset, and the rest $28$ are used to form the data pool for active learning. The $60$ drilling locations provide $5186$ labeled operational data, which are split   
into two parts: $3957$ of them are used for training and the rest $1229$ are used for testing. In the initial training stage, $75\%$ of the training data are used to update the weights and the rest $25\%$ are used for the validation of hyper-parameters.

For each one of the $28$ drilling locations, we choose five operational data that lie at one-quarter, midpoint, three-quarter and both ends of the 0.6m-long line segment centering on the drilling location. The resultant $28 \times 5$ operational data form the data pool ${\cal P}$ for active learning. As long as an operational datum in ${\cal P}$ is recommended by the active learning method, we retrieve the drilling location corresponding to the operational datum and set the labeled operational data within $0.3$m around this drilling location as the newly-added training samples.

\subsection{Experiment Setting}

In the feature-extraction part of iGAN-GR's generator, the MSA module has $L=8$ heads, and no trainable weights are set in the PE module. The fully-connected part of iGAN-GR's generator has a 20-7-11 structure and symmetrically, iGAN-GR's discriminator has a structure of 11-7-20-2 fully-connected layers. The activation functions of the input and the hidden nodes of the two sub-networks are set as the ReLU functions. The output nodes of  generator are activated by using the softmax function and the two output nodes of discriminator are both activated by using the Sigmoid function, respectively. The hyper-parameters of Adam optimizer for training iGAN-GR are listed in Tab. \ref{tab:adam}

\begin{table*}[htbp]
\centering \vskip -0.1in
	\caption{The choices of hyper-parameters in the training of iGAN-GR}\label{tab:adam}
	\resizebox{0.8\linewidth}{!}{
	\begin{threeparttable}
	\begin{tabular}{l l l l l l l l l l}
		\hline
		Stage        &  $ K $  & $ \eta_g $ &$ \eta_d $ &$ \eta_s $ & Size & $\lambda_d$ & $\lambda_g$ &$\beta_d$ &$\beta_g$\\ \hline
	Pre-training         & 200     & -          & -         & 0.005     & 32 	& - 		  & -           & -        & -  \\ \hline 
	Initial training    & 300     & 0.0001     & 0.0005    & 0.001     & 88 	& -           & -           & -        & -  \\ \hline 
	Incremental training & 100     & 0.0001     & 0.0005    & 0.001     & 32 	& 500         & 500         & 1        & 1  \\ \hline 
	\end{tabular}
	\begin{tablenotes}
				\item[1] ``K"  is the number of training iterations.
				\item[2] $ \eta_g $, $ \eta_d $ and $ \eta_s$ are the learning rates for minimizing the generative loss ${\cal L}_{\rm G}$, the discriminator loss ${\cal L}_{\rm D}$ and the loss ${\cal L}_{\rm S}$, respectively.
				\item[3] ``Size" stands for the size of minibatch in the Adam optimizer.
				\item[4] $\lambda_d$ and $\lambda_g$ (resp. $\beta_d$ and $\beta_g$) are the control factors corresponding to the EWC (resp. IS) loss, respectively.
	\end{tablenotes}
		\end{threeparttable}
	} \vskip -0.1in
\end{table*}

Given a testing dataset $\overline{\mathcal{S}}:=\left\{\overline{\textbf{x}}_n,\overline{{\bf y}}_n\right\}_{n=1}^{N}$, we adopt the mean square error (MSE) over five repeated experiments on it as the performance criterion:
\begin{equation*}\label{equ:mse}
	{\rm MSE}= \frac{1}{N}\sum_{n=1}^{N}\left[\overline{{\bf y}}_n-{\rm G}[\overline{\textbf{x}}_n]\right]^2
\end{equation*}
where ${\rm G}[\overline{\textbf{x}}_n]$ is the output of the generator w.r.t. the testing input $\overline{\textbf{x}}_n$. Additionally, we adopt four state-of-the-art regressors as the comparative models including random forest regression (RFR), support vector regression (SVR), Xgboost regressor (XGBR) and GAN-GP \cite{zhang2022generative}. The hyper-parameters of the first three models are determined by using five-fold cross validation and those of GAN-GP follow the relevant suggestion in \cite{zhang2022generative}. For the sake of fairness, after querying samples every time, these comparative models share the same samples that are used to incrementally training iGAN-GR.

We adopt two kinds of active learning methods including the EUS method and the QBC method. Either of the two methods is repeatedly implemented nine times to query candidate operational data from the data pool. We introduce the performance-gain rate $ \gamma $ to measure the influence caused by the active learning methods to the reconstruction performance:
\begin{equation}
	\gamma_t = \frac{{\rm MSE}_{t-1}-{\rm MSE}_{t}}{{\rm MSE}_{t-1}}, \quad t\geq 1,
\end{equation}
where ${\rm MSE}_t$ stands for the MSE of the reconstruction model after the $t$-th sample querying of active learning. In the QBC method, the committee models are set as ten iGAN-GRs, each of which is trained via the aforementioned workflow ({\it cf.} Alg. \ref{alg:al-igan}) based on randomly-selecting $80\%$ of the original training data. It is noteworthy that, in the QBC method for each comparative model, the committee models are set as the same models that are trained in the aforementioned way. As a comparison, we also consider the random sampling (RS) method that randomly selects a point from the data pool and the newly-added training data are the labeled operational data associated with its corresponding drilling location.

\subsection{Experimental Results}

\begin{table}[htbp]
	\caption{Averaged reconstruction performance (MSE) of different models with active learning methods}
	\label{tab:mse}
	\centering
	\resizebox{0.7\linewidth}{!}{
		\begin{threeparttable} 
			\begin{tabular}{c|c|ccccc}
				\hline
				T& \diagbox{AL}{Model} & SVR & RFR & XGBR & GAN-GP & iGAN-GR\\
				\hline
				0 & RS  & 0.0187 & 0.0202 & 0.0191 & 0.0224 & 0.0190  \\ 
				~ & EUS & 0.0187 & 0.0205 & 0.0191 & 0.0206 & 0.0189  \\ 
				~ & QBC & 0.0187 & 0.0213 & 0.0191 & 0.0190 & {\bf 0.0183}  \\ 
				\hline
				1 & RS  & 0.0181 & 0.0183 & 0.0183 & 0.0183 & 0.0181  \\ 
				~ & EUS & 0.0172 & 0.0199 & 0.0186 & 0.0171 & 0.0182  \\ 
				~ & QBC & 0.0182 & 0.0185 & 0.0186 & 0.0192 & {\bf0.0168}  \\ 
				\hline
				2 & RS  & 0.0176 & 0.0168 & 0.0178 & 0.0179 & 0.0159  \\ 
				~ & EUS & 0.0161 & 0.0173 & 0.0183 & 0.0194 & 0.0175  \\ 
				~ & QBC & 0.0177 & 0.0178 & 0.0161 & 0.0180 & {\bf0.0156}  \\ 
				\hline
				3 & RS  & 0.0172 & 0.0172 & 0.0192 & 0.0189 & 0.0157  \\ 
				~ & EUS & 0.0154 & 0.0179 & 0.0168 & 0.0187 & 0.0165  \\ 
				~ & QBC & 0.0165 & 0.0186 & 0.0167 & 0.0164 & {\bf0.0149}  \\ 
				\hline
				4 & RS  & 0.0169 & 0.0182 & 0.0200 & 0.0175 & 0.0145  \\ 
				~ & EUS & 0.0150 & 0.0174 & 0.0165 & 0.0182 & 0.0162  \\ 
				~ & QBC & 0.0163 & 0.0181 & 0.0163 & 0.0173 & {\bf0.0142}  \\ 
				\hline
				5 & RS  & 0.0166 & 0.0188 & 0.0160 & 0.0170 & 0.0145  \\ 
				~ & EUS & 0.0147 & 0.0200 & 0.0161 & 0.0149 & 0.0148  \\ 
				~ & QBC & 0.0160 & 0.0188 & 0.0167 & 0.0190 & {\bf 0.0142}  \\ 
				\hline
				6 & RS  & 0.0164 & 0.0195 & 0.0165 & 0.0248 & 0.0142  \\ 
				~ & EUS & 0.0146 & 0.0207 & 0.0184 & 0.0180 & {\bf 0.0126}  \\ 
				~ & QBC & 0.0158 & 0.0217 & 0.0168 & 0.0210 & 0.0138  \\ 
				\hline
				7 & RS  & 0.0160 & 0.0215 & 0.0156 & 0.0195 & 0.0134  \\ 
				~ & EUS & 0.0146 & 0.0225 & 0.0177 & 0.0159 & {\bf 0.0115}  \\ 
				~ & QBC & 0.0161 & 0.0234 & 0.0150 & 0.0201 & 0.0131  \\ 
				\hline
				8 & RS  & 0.0162 & 0.0250 & 0.0165 & 0.0236 & 0.0118  \\ 
				~ & EUS & 0.0148 & 0.0260 & 0.0184 & 0.0165 & {\bf 0.0110}  \\ 
				~ & QBC & 0.0163 & 0.0263 & 0.0158 & 0.0248 & 0.0127  \\ 
				\hline
				9 & RS  & 0.0164 & 0.0265 & 0.0164 & 0.0243 & 0.0112  \\ 
				~ & EUS & 0.0152 & 0.0285 & 0.0174 & 0.0165 & {\bf 0.0101}  \\ 
				~ & QBC & 0.0167 & 0.0298 & 0.0173 & 0.0209 & 0.0104  \\ 
				\hline
			\end{tabular}
			\begin{tablenotes}
				\item[1] T is the number of times of querying samples.
				\item[2] AL is the abbreviation of active learning. 
					\end{tablenotes}
		\end{threeparttable}
	}

\end{table}

\begin{figure*}[htbp]
	\centering
\subfloat[MSE for RS]{
	\begin{minipage}[t]{0.33\linewidth}
		\centering
		\includegraphics[width=\linewidth]{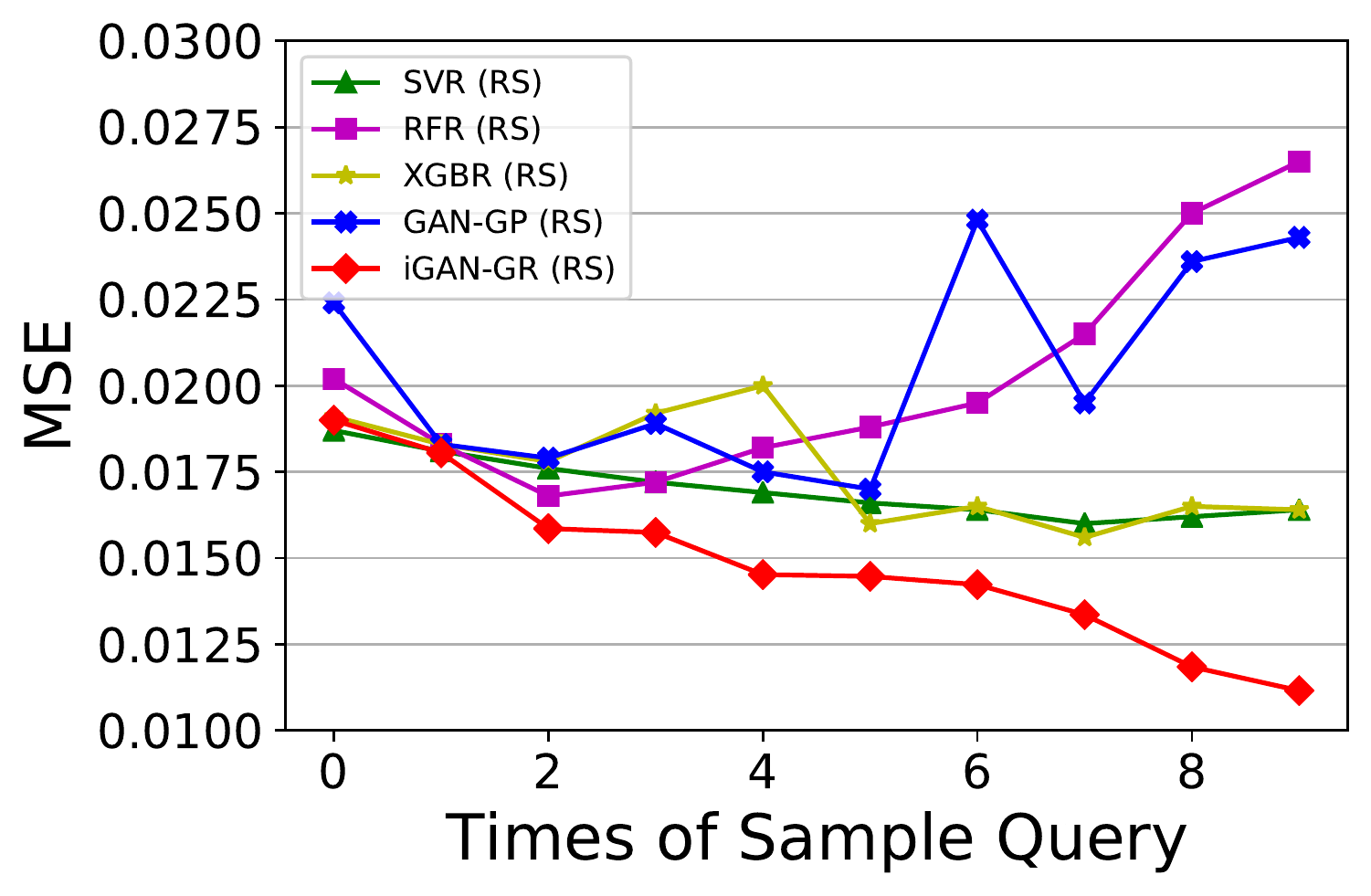}
	\end{minipage}%
}
\subfloat[MSE for EUS]{
	\begin{minipage}[t]{0.33\linewidth}
		\centering
		\includegraphics[width=\linewidth]{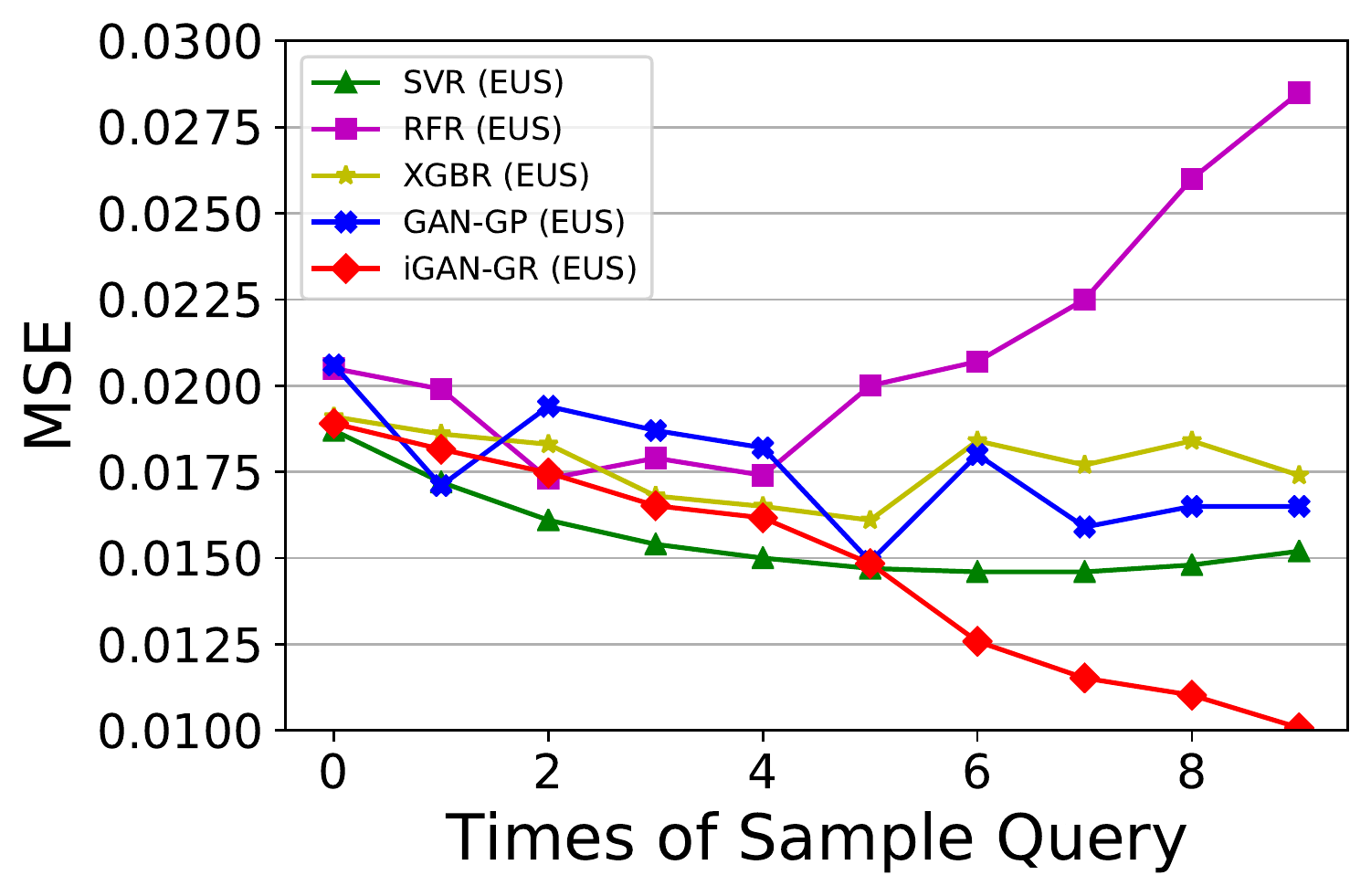}
	\end{minipage}%
}
\subfloat[MSE for QBC]{
	\begin{minipage}[t]{0.33\linewidth}
		\centering
		\includegraphics[width=\linewidth]{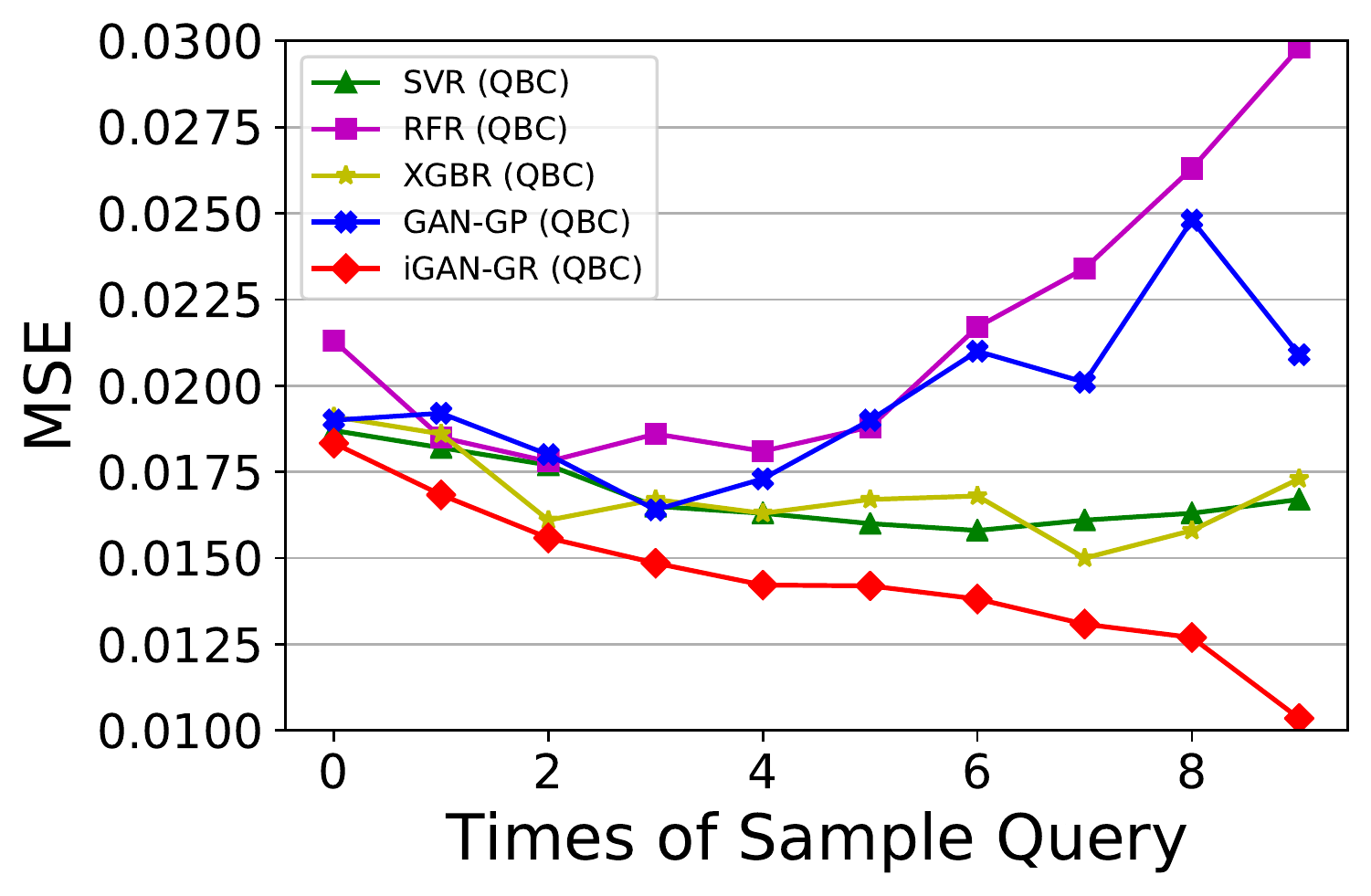}
	\end{minipage}%
}
\centering	
	\caption{Geological reconstruction performance of different models with active learning methods.}
	\label{fig:mse}
\end{figure*}

As shown in Fig. \ref{fig:mse} and Tab. \ref{tab:mse}, taking advantages of the specific structure of discriminator and the appropriate incremental training strategy, the proposed iGAN-GR outperforms the other models in the task of geological reconstruction with incremental training samples that are recommended by active learning. The existence of discriminator output ${\rm D}_{\rm o/f}[\cdot]$ makes iGAN-GR prevent from forgetting the previous training results when the newly-added samples are available for incrementally updating the weights of iGAN-GR. In contrast, although GAN-GP has the almost same structure of the proposed iGAN-GR except the discriminator output ${\rm D}_{\rm o/f}[\cdot]$ and is also trained by using the same samples, its reconstruction performance becomes worse when the number of sample query increases. This phenomenon validates iGAN-GR's the specific structure and training strategy for incremental learning. Moreover, the EUS and the QBC active learning methods perform better than the RS method in the incremental training stage of iGAN-GR

\begin{table}[htbp]
	\caption{ Performance gain $ \gamma_t $ of different models after the $t$-th active learning }
	\label{tab:gamma}
	\centering
	\resizebox{\linewidth}{!}{
		\begin{threeparttable} 
			\begin{tabular}{c|ccccc|ccccc|ccccc}
				\hline
				AL  & &  &RS& &   & & &  EUS  & & &  & &QBC &  & \\  
				\hline
				\diagbox{T}{Model}&SVR & RFR & XGBR & GAN-GP & iGAN-GR & SVR & RFR & XGBR & GAN-GP & iGAN-GR &SVR & RFR & XGBR & GAN-GP & iGAN-GR\\  
				\hline
				1 & 0.0321 & 0.0941 & 0.0419 & 0.1830 & 0.0497 & 0.0802 & 0.0293 & 0.0262 & 0.1699 & 0.0395 & 0.0267 & 0.1315 & 0.0262 & \colorbox{mygray}{-0.0105} & 0.0817  \\ 
		        2 & 0.0276 & 0.0820 & 0.0273 & 0.0219 & 0.1217 & 0.0640 & 0.1307 & 0.0161 & \colorbox{mygray}{-0.1345} & 0.0376 & 0.0275 & 0.0378 & 0.1344 & 0.0625 & 0.0743  \\ 
		        3 & 0.0227 & \colorbox{mygray}{-0.0238} & \colorbox{mygray}{-0.0787} & \colorbox{mygray}{-0.0559} & 0.0071 & 0.0435 & \colorbox{mygray}{-0.0347} & 0.0820 & 0.0361 & 0.0546 & 0.0678 & \colorbox{mygray}{-0.0449} & \colorbox{mygray}{-0.0373} & 0.0889 & 0.0469  \\ 
		        4 & 0.0174 & \colorbox{mygray}{-0.0581} & \colorbox{mygray}{-0.0417} & 0.0741 & 0.0776 & 0.0260 & 0.0279 & 0.0179 & 0.0267 & 0.0213 & 0.0121 & 0.0269 & 0.0240 & \colorbox{mygray}{-0.0549} & 0.0428  \\ 
		        5 & 0.0178 & \colorbox{mygray}{-0.0330} & 0.2000 & 0.0286 & 0.0036 & 0.0200 & \colorbox{mygray}{-0.1494} & 0.0242 & 0.1813 & 0.0820 & 0.0184 & \colorbox{mygray}{-0.0387} & \colorbox{mygray}{-0.0245} & \colorbox{mygray}{-0.0983} & 0.0017  \\ 
		        6 & 0.0120 & \colorbox{mygray}{-0.0372} & \colorbox{mygray}{-0.0313} & \colorbox{mygray}{-0.4588} & 0.0166 & 0.0068 & \colorbox{mygray}{-0.0350} & \colorbox{mygray}{-0.1429} & \colorbox{mygray}{-0.2081} & 0.1521 & 0.0125 & \colorbox{mygray}{-0.1543} & \colorbox{mygray}{-0.0060} & \colorbox{mygray}{-0.1053} & 0.0268  \\ 
		        7 & 0.0244 & \colorbox{mygray}{-0.1026} & 0.0545 & 0.2137 & 0.0613 & 0.0000 & \colorbox{mygray}{-0.0870} & 0.0380 & 0.1167 & 0.0845 & \colorbox{mygray}{-0.0190} & \colorbox{mygray}{-0.0783} & 0.1071 & 0.0429 & 0.0529  \\ 
		        8 & \colorbox{mygray}{-0.0125} & \colorbox{mygray}{-0.1628} & \colorbox{mygray}{-0.0577} & \colorbox{mygray}{-0.2103} & 0.1132 & \colorbox{mygray}{-0.0137} & \colorbox{mygray}{-0.1556} & \colorbox{mygray}{-0.0395} & \colorbox{mygray}{-0.0377} & 0.0430 & \colorbox{mygray}{-0.0124} & \colorbox{mygray}{-0.1239} & \colorbox{mygray}{-0.0533} & \colorbox{mygray}{-0.2338} & 0.0295  \\ 
		        9 & \colorbox{mygray}{-0.0123} & \colorbox{mygray}{-0.0600} & 0.0061 & \colorbox{mygray}{-0.0297} & 0.0581 & \colorbox{mygray}{-0.0270} &\colorbox{mygray}{ -0.0962} & 0.0543 & 0.0000 & 0.0858 & \colorbox{mygray}{-0.0245} & \colorbox{mygray}{-0.0331} & \colorbox{mygray}{-0.0949} & 0.1573 & 0.1848  \\ 
				\hline
			\end{tabular}
			\begin{tablenotes}
				\item[1] T is the number of times of querying samples.
				\item[2] AL is the abbreviation of active learning . 
					\end{tablenotes}
		\end{threeparttable}
	}
\end{table}

\begin{figure*}[htbp]
	\centering
	\subfloat[$ \gamma_t $ for RS]{
		\begin{minipage}[t]{0.33\linewidth}
			\centering
			\includegraphics[width=\linewidth]{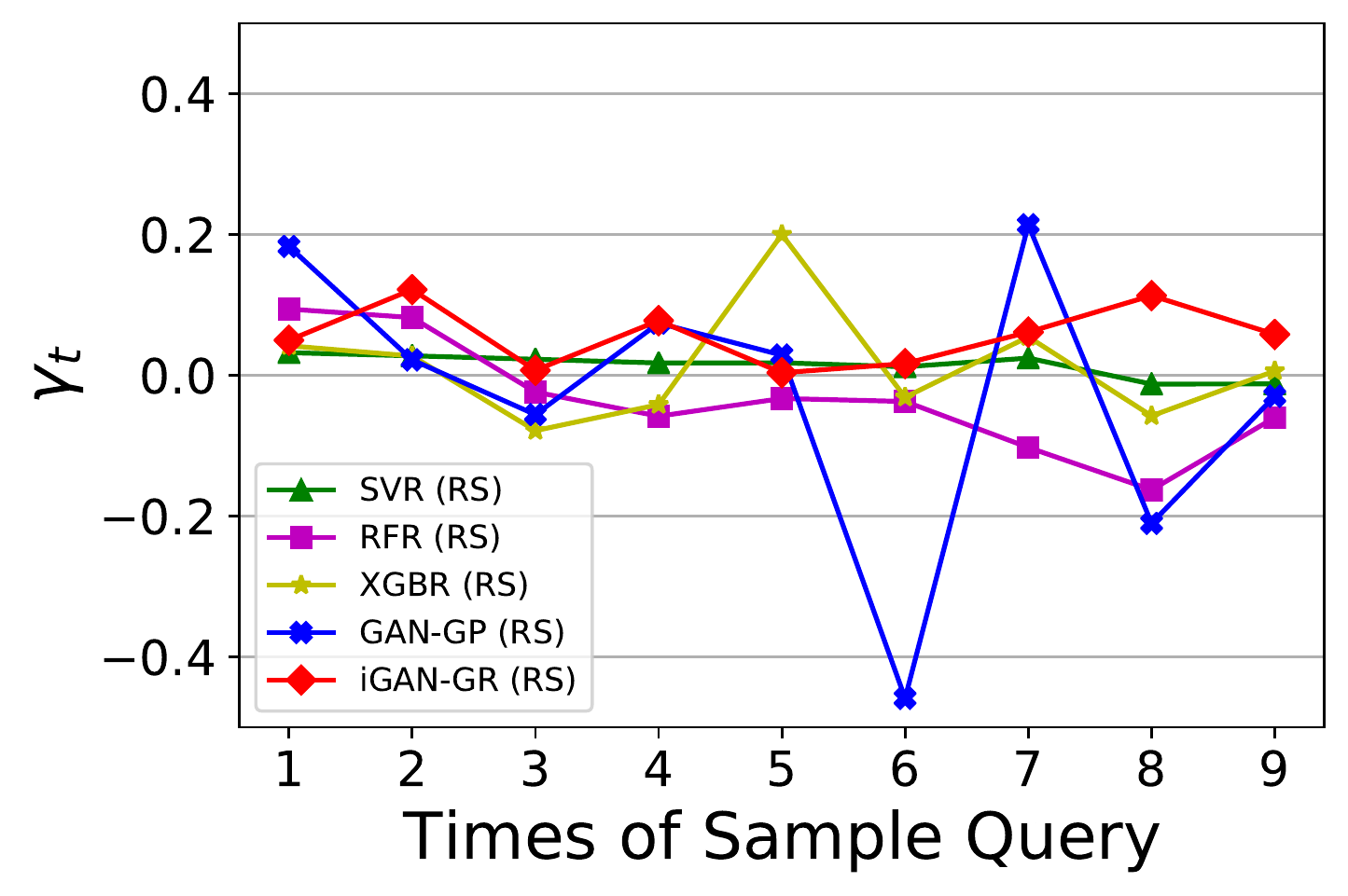}
		\end{minipage}%
	}
	\subfloat[$ \gamma_t $ for EUS]{
		\begin{minipage}[t]{0.33\linewidth}
			\centering
			\includegraphics[width=\linewidth]{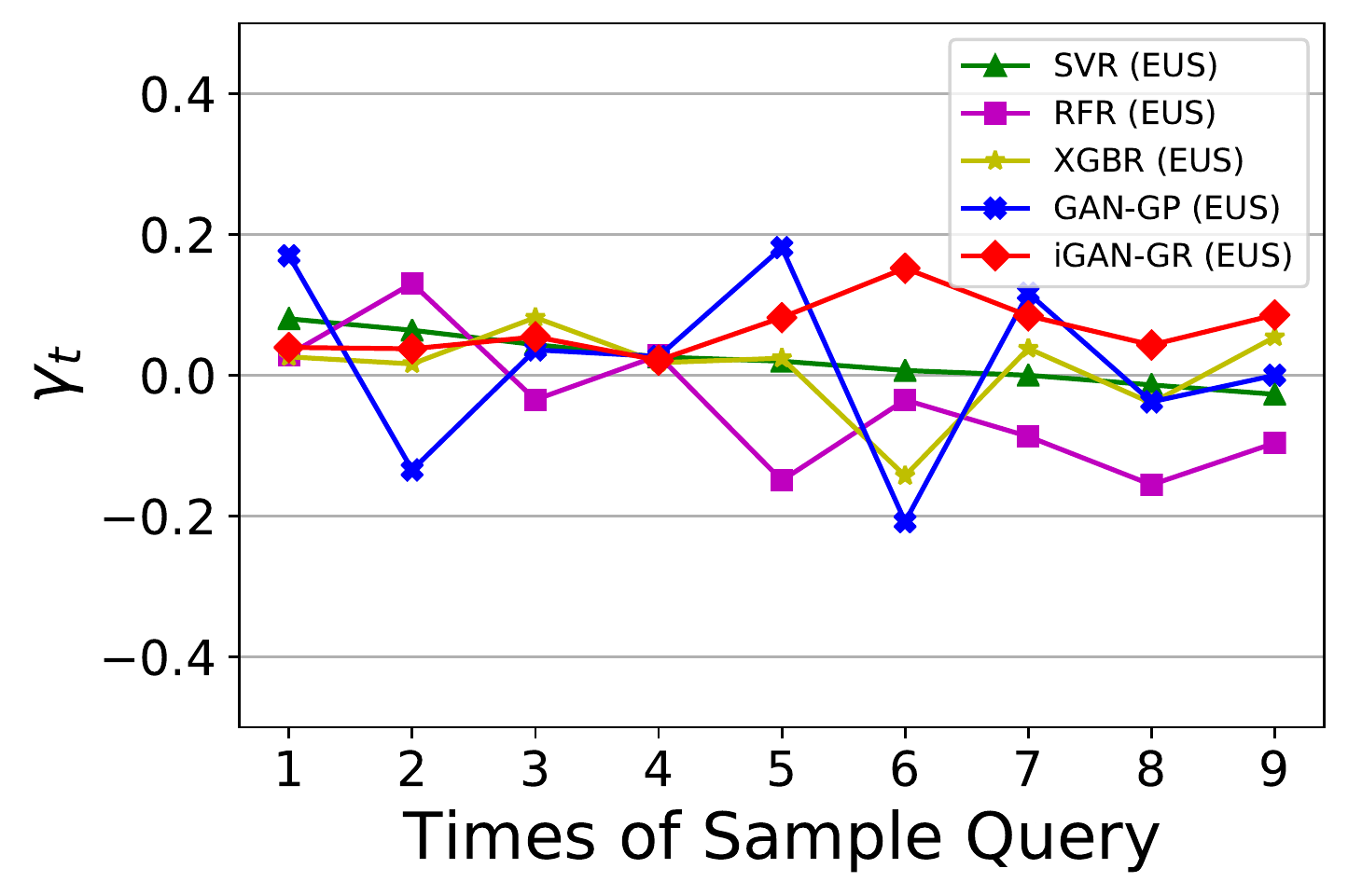}
		\end{minipage}%
	}
	\subfloat[$ \gamma_t $ for QBC]{
		\begin{minipage}[t]{0.33\linewidth}
			\centering
			\includegraphics[width=\linewidth]{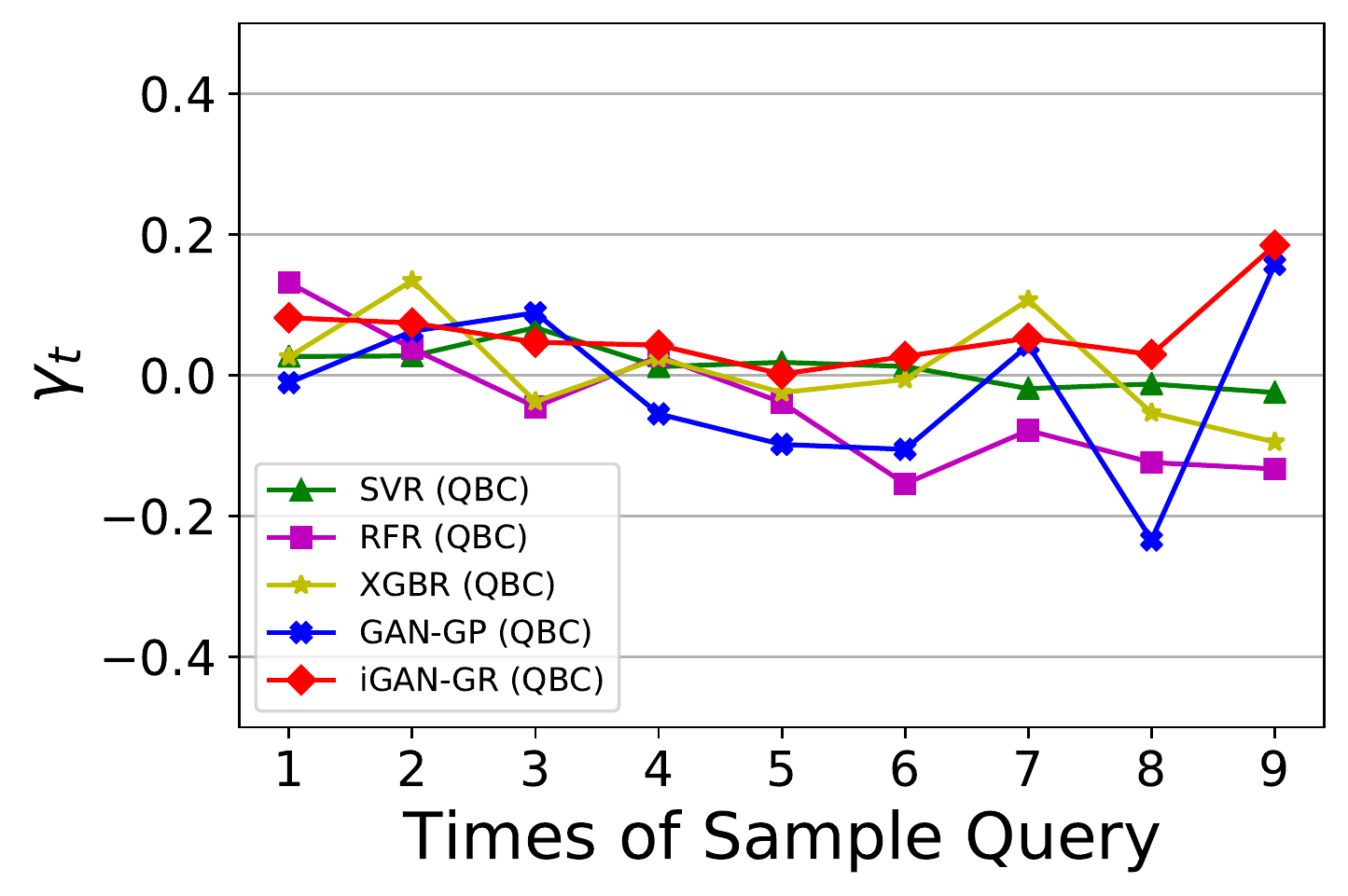}
		\end{minipage}%
	}
	\centering	
	\caption{Performance gain $ \gamma_t $ of different models after querying samples via active learning.}
	\label{fig:gamma}
\end{figure*}

Moreover, we also examine the performance gain provided by the newly-added training samples recommended by using active learning methods. As shown in Fig. \ref{fig:gamma} and Tab. \ref{tab:gamma}, after each time of sample querying, the iGAN-GR has a stable and positive performance gain, and in contrast, the performance gains of other models are not always positive and have many fluctuations as the number of times of querying samples increases. This experiment phenomenon also verify the effectiveness of the proposed iGAN-GR. 

Because of the complicated network structure, the time cost of training iGAN-GR (and GAN-GP) is higher than the other models, and especially, the QBC active learning methods have a extremely high time cost because there are nine committee members to be trained at each time of querying samples ({\it cf.} Tab. \ref{tab:cost}). However, the testing time cost of iGAN-GR keeps at a reasonable level, and in contrast, SVR is time-consuming in the testing phase because each testing output is computed by traversing all support vectors ({\it i.e.,} training samples). 

To sum up, the experiment demonstrates the proposed iGAN-GR with the specific training strategy, which contains two stages: initial training and incremental training, can successfully the task of geological reconstruction, and has a significant superiority to the other models. The EUS and the QBC active learning methods both perform well, but the former has a better feasibility in practice because of its much lower time cost.

\begin{table}[htbp]
	\caption{Time cost of different models with active learning methods}
	\label{tab:cost}
	\centering
	\resizebox{\linewidth}{!}{
		\begin{threeparttable} 
			\begin{tabular}{c|c|ccccc|ccccc|ccccc}
				\hline
				 &AL  & &  &RS& &   & & &  EUS  & & &  & &QBC &  & \\  
				\hline
				T&\diagbox{Phase}{Model}&SVR & RFR & XGBR & GAN-GP & iGAN-GR & SVR & RFR & XGBR & GAN-GP & iGAN-GR &SVR & RFR & XGBR & GAN-GP & iGAN-GR\\  
				\hline
				0 & training (s) & 4.0845 & 68.0003 & 18.0856 & 230.0081 & 184.8574 & 4.1455 & 68.1935 & 18.1435 & 211.5 & 333.943 & 4.026 & 71.8587 & 18.1315 & 210.0999 & 185.9665  \\
				~ & testing (s) & 1.3948 & 0.007 & 0.0862 & 0.0018 & 0.0014 & 1.3791 & 0.007 & 0.0884 & 0.0018 & 0.003 & 1.3681 & 0.0068 & 0.0902 & 0.0018 & 0.0016  \\ 
				~ & querying (s) & 0.0002 & 0.0000 & 0.0002 & 0.0044 & 0.0058 & 1.8463 & 0.0708 & 0.8046 & 0.3183 & 0.3912 & 42.0542 & 57.6359 & 156.9674 & 2741.5014 & 2409.3793  \\ \hline
				1 & training (s) & 3.1443 & 60.1839 & 16.2455 & 171.6557 & 154.6756 & 2.5316 & 63.1164 & 15.6661 & 173.3605 & 274.8515 & 3.2691 & 61.9104 & 15.8286 & 175.2347 & 152.5634  \\ 
				~ & testing (s) & 1.2837 & 0.0066 & 0.0878 & 0.0024 & 0.0016 & 1.2577 & 0.0074 & 0.0906 & 0.0018 & 0.0036 & 1.3055 & 0.0066 & 0.0908 & 0.0014 & 0.0018  \\ 
				~ & querying (s) & 0.0002 & 0.0002 & 0.0002 & 0.005 & 0.0056 & 1.6298 & 0.0792 & 0.7797 & 0.311 & 0.3726 & 36.98 & 43.3032 & 133.2579 & 2006.1148 & 2002.8516  \\ \hline
				2 & training (s) & 2.1391 & 49.7725 & 13.8729 & 151.3236 & 129.1286 & 2.2568 & 49.2805 & 13.3573 & 152.8024 & 180.1242 & 2.8913 & 46.8407 & 13.4121 & 152.918 & 128.8395  \\ 
				~ & testing (s) & 1.1985 & 0.009 & 0.088 & 0.0032 & 0.0014 & 1.1936 & 0.0067 & 0.0876 & 0.0014 & 0.0026 & 1.2106 & 0.0072 & 0.0898 & 0.0012 & 0.0012  \\ 
				~ & querying (s) & 0.0000 & 0.0004 & 0.0006 & 0.0048 & 0.0048 & 1.4613 & 0.0662 & 0.7476 & 0.2926 & 0.3187 & 30.6326 & 37.5006 & 114.5639 & 1738.0138 & 1676.7065  \\ \hline
				3 & training (s) & 1.6088 & 42.2052 & 11.4274 & 126.4904 & 107.6348 & 1.4284 & 38.6755 & 11.6239 & 124.1489 & 102.3696 & 2.0797 & 38.4342 & 11.3756 & 134.5831 & 118.9602  \\ 
				~ & testing (s) & 1.0666 & 0.007 & 0.0886 & 0.0032 & 0.0014 & 1.0982 & 0.0068 & 0.09 & 0.0014 & 0.0014 & 1.136 & 0.0071 & 0.0894 & 0.0016 & 0.0018  \\ 
				~ & querying (s) & 0.0002 & 0.0001 & 0.0004 & 0.0036 & 0.005 & 1.2477 & 0.063 & 0.7125 & 0.2886 & 0.2868 & 24.9794 & 29.611 & 96.9935 & 1462.4156 & 1539.5195  \\ \hline
				4 & training (s) & 1.1754 & 35.0135 & 9.7389 & 93.846 & 90.2674 & 1.0738 & 31.3401 & 9.726 & 99.4397 & 85.08 & 1.3197 & 31.8418 & 9.8546 & 106.5846 & 89.3011  \\ 
				~ & testing (s) & 0.9818 & 0.0068 & 0.0874 & 0.003 & 0.0014 & 1.0118 & 0.0066 & 0.0908 & 0.0018 & 0.0018 & 1.0126 & 0.0066 & 0.09 & 0.0023 & 0.0018  \\
				~ & querying (s) & 0.0004 & 0.0004 & 0.0002 & 0.0034 & 0.0044 & 1.0836 & 0.0619 & 0.6854 & 0.2759 & 0.2715 & 18.469 & 23.3716 & 83.6602 & 1227.1505 & 1164.9315  \\ \hline
				5 & training (s) & 0.8108 & 28.3352 & 8.8772 & 75.875 & 76.5524 & 0.7674 & 29.2835 & 8.6159 & 76.4178 & 71.0697 & 0.8364 & 26.4632 & 8.9291 & 79.399 & 69.5592  \\ 
				~ & testing (s) & 0.8917 & 0.0068 & 0.0906 & 0.0032 & 0.0014 & 0.9215 & 0.007 & 0.0886 & 0.0016 & 0.0016 & 0.9217 & 0.0072 & 0.0888 & 0.0014 & 0.0016  \\
				~ & querying (s) & 0.0002 & 0.0004 & 0.0000 & 0.0036 & 0.0044 & 0.9537 & 0.0642 & 0.6467 & 0.2559 & 0.2603 & 14.8358 & 18.2467 & 71.4518 & 918.4933 & 924.0885  \\ \hline
				6 & training (s) & 0.6781 & 24.0696 & 7.5957 & 65.0571 & 65.876 & 0.648 & 22.1098 & 7.2326 & 69.3814 & 60.5286 & 0.6177 & 23.163 & 7.0491 & 65.6143 & 58.3781  \\ 
				~ & testing (s) & 0.8391 & 0.0068 & 0.089 & 0.0036 & 0.0017 & 0.8823 & 0.007 & 0.0896 & 0.0016 & 0.0016 & 0.8418 & 0.0068 & 0.088 & 0.0014 & 0.0016  \\ \
				~ & querying (s) & 0.0004 & 0.0002 & 0.0002 & 0.0038 & 0.0042 & 0.8741 & 0.0581 & 0.6213 & 0.2475 & 0.2426 & 12.3204 & 15.1267 & 61.3728 & 756.9021 & 765.4794  \\ \hline
				7 & training (s) & 0.4915 & 20.5262 & 6.3739 & 55.3658 & 56.487 & 0.4639 & 19.6195 & 6.1047 & 49.2135 & 51.2222 & 0.4211 & 18.3329 & 6.1745 & 47.3037 & 49.1769  \\ 
				~ & testing (s) & 0.7606 & 0.0074 & 0.0868 & 0.0026 & 0.0012 & 0.8021 & 0.0072 & 0.089 & 0.0016 & 0.0014 & 0.758 & 0.0064 & 0.0896 & 0.002 & 0.0014  \\ 
				~ & querying (s) & 0.0004 & 0.0004 & 0.0004 & 0.0024 & 0.0034 & 0.7552 & 0.0553 & 0.5894 & 0.2368 & 0.2348 & 9.8574 & 12.3273 & 52.4706 & 639.6258 & 645.2317  \\ \hline
				8 & training (s) & 0.3734 & 17.4084 & 5.3307 & 47.7936 & 48.345 & 0.3643 & 18.4755 & 5.2789 & 43.5528 & 44.9365 & 0.3571 & 15.4815 & 5.3826 & 41.5932 & 40.8628  \\ 
				~ & testing (s) & 0.6923 & 0.0062 & 0.0886 & 0.0028 & 0.0014 & 0.7529 & 0.007 & 0.086 & 0.0014 & 0.0012 & 0.7037 & 0.0066 & 0.0894 & 0.0014 & 0.0018  \\ 
				~ & querying (s) & 0.0002 & 0.0000 & 0.0008 & 0.003 & 0.0032 & 0.6634 & 0.0534 & 0.5623 & 0.2188 & 0.2254 & 8.3672 & 10.2366 & 46.2685 & 541.9268 & 555.6884  \\ \hline
				9 & training (s) & 0.2895 & 16.0201 & 4.6685 & 39.2488 & 42.5663 & 0.2793 & 16.5178 & 4.4172 & 38.3525 & 39.0722 & 0.271 & 13.9656 & 4.474 & 37.6931 & 36.3782  \\ 
				~ & testing (s) & 0.6241 & 0.0064 & 0.0888 & 0.0032 & 0.0018 & 0.6455 & 0.0074 & 0.0904 & 0.0014 & 0.0014 & 0.6276 & 0.0068 & 0.089 & 0.0018 & 0.0014  \\ 
				~ & querying (s) & 0.0000 & 0.0002 & 0.0002 & 0.0022 & 0.003 & 0.5421 & 0.0534 & 0.5304 & 0.215 & 0.2076 & 7.1891 & 8.3745 & 40.658 & 445.6629 & 460.7213  \\ \hline
			\end{tabular}
		\end{threeparttable}
	}
\end{table}

\section{Conclusion}

In this paper, we propose a framework, called AL-iGAN for TBM tunnel geological reconstruction. This framework contains three aspects: 1) the active learning method for selecting the most informative fresh samples to improve the current model; 2) the iGAN-GR model to estimate the thickness of each kind of rock-soil type at a location of the tunnel based on TBM operational data indexed by continuous displacement; and 3) the training strategy designed for incremental learning.

We adopt the EUS and the QBC active learning methods to select a candidate operational datum from the data pool, and then retrieve the location corresponding to the selected datum. Then, the geological information is obtained by taking a drilling at this location, and then the operational data within 0.3m around the location are labeled with the thickness of the observed rock-soil types to form the newly-added training samples. The numerical experiments show that the two kinds of active learning methods perform comparably, but the time cost of the QBC method is much higher than that of the EUS method. Therefore, the EUC method could be an appropriate choice in practice. It is noteworthy that being not limited to the drilling methods, there have been other ways to detect the geological information at one individual location, for example, seismic ahead prospecting methods \cite{liu2018new,liu2019comprehensive}, electromagnetic ahead prospecting methods \cite{li2022new,liu2018forward} and electrical and induced polarization ahead prospecting methods\cite{belova2021mineral,alfouzan2020spectral}.

The structure of iGAN-GR is specifically designed for the incremental learning of the TBM tunnel geological reconstruction based on TBM operational data. In view of the strong sequentiality of these data, we combine the MSA module with the PE module to form the feature-extraction part in the generator of iGAN-GR. We also impose two output nodes ${\rm D}_{\rm r / g}[\cdot]$ and ${\rm D}_{\rm o / f}[\cdot]$ in the discriminator of iGAN-GR: ${\rm D}_{\rm r / g}[\cdot]$ is to identify whether the input instance is a real training sample or the output of generator; and ${\rm D}_{\rm r / g}[\cdot]$ is to justify whether the input instance is one of the original or the newly-added training samples. 

The training of iGAN-GR is achieved via two stages: initial training and incremental training. First, we use the original training samples to update the weights of an iGAN-GR ({\it cf.} Alg. \ref{alg:initial}); and then we use the new training samples, a part of which are the newly-added training samples, to refine the weights of the iGAN-GR for fitting the fresh samples ({\it cf.} Alg. \ref{alg:incremental}). The usage of ${\rm D}_{\rm r / g}[\cdot]$ avoids the previously-trained iGAN-GR to forget the previous memory in the incremental training stage.

The collaboration of the three aspects results in the proposed AL-iGAN framework ({\it cf.} Alg. \ref{alg:al-igan}). The experimental results support the effectiveness of the AL-iGAN framework in the task of incremental geological reconstruction and show that 1) iGAN-GR outperforms the other models when only some of original training samples are available; and 2) taking advantages of specific structure and training strategy, iGAN-GR has the best reconstruction accuracy with a positive performance gain after each time of querying samples in active learning. In future works, we will consider the applications of the diffusion models and reinforcement learning in the task of incremental geological reconstruction.

\begin{table}[htbp]
	\caption{Physical and Mechanical Indicators of Different rock-soil types and the number of times that each rock-soil type appears in the drilling samples}\label{tab:rock-soil}
	\centering
	\resizebox{120mm}{!}{
		\begin{threeparttable} 
			\begin{tabular}{l|ccccccc|c}
				\toprule
				\multicolumn{1}{c|}{\diagbox{Type}{Indicator}} & Y (${\rm kN/m^3}$) & FI ($^\circ $) & EM (${\rm MPa}$) & P & SITA & K (${\rm m/d}$) & FRB (${\rm kPa}$) & Number \\ 
				\midrule
				$\large{\textcircled{\footnotesize{2}}}_3$      & 17.000 & 4.500  & 4.000 & 0.400 & 0.650 & 0.003 &10.000 & 7    \\
				$\large{\textcircled{\footnotesize{4}}}_2$       & 19.000 & 15.000  & 15.000 & 0.320 & 0.500 & 0.005 &25.000 & 3    \\
				$\large{\textcircled{\footnotesize{4}}}_4$       & 18.000 & 8.000  & 4.500 & 0.420 & 0.700 & 0.005 &18.000 & 2    \\
				$\large{\textcircled{\footnotesize{4}}}_{10}$      & 20.500 & 32.000  & 25.000 & 0.220 & 0.350 & 20.000 &55.000 & 9    \\
				$\large{\textcircled{\footnotesize{7}}}_{2-1}$   & 18.500 & 20.500  & 18.000 & 0.300 & 0.450 & 0.500 &22.000 & 9    \\
				$\large{\textcircled{\footnotesize{7}}}_{2-2}$   & 18.500 & 22.500  & 20.000 & 0.280 & 0.550 & 0.500 &28.000 & 36   \\
				$\large{\textcircled{\footnotesize{9}}}_1$        & 19.500 & 25.000  & 40.000 & 0.250 & 0.000 & 0.800 &45.000 & 9    \\
				$\large{\textcircled{\footnotesize{9}}}_{2-1}$   & 20.500 & 27.500  & 90.000 & 0.250 & 0.000 & 2.500 &60.000 & 5    \\
				$\large{\textcircled{\footnotesize{12}}}_{1}$    & 19.500 & 27.000  & 40.000 & 0.250 & 0.000 & 1.000 &45.000 & 6    \\
				$\large{\textcircled{\footnotesize{12}}}_{2-1}$  & 20.500 & 30.000  & 90.000 & 0.250 & 0.000 & 2.500 &60.000 & 1    \\
				$\large{\textcircled{\footnotesize{12}}}_{3}$    & 24.500 & 55.000  & 10000 & 0.220 & 0.000 & 1.500 &380.000 & 1    \\
				\bottomrule
			\end{tabular}           
		\end{threeparttable}
	}
\end{table}

\section{Acknowledgement}
This work is supported by the National Key R\&D Program of China (No. 2018YFB1702502) and the National Natural Science Foundation of China (No. 62176040 and No. 52075068)

\bibliography{AL-iGAN}

\end{document}